\documentclass[%
reprint,
superscriptaddress,
 amsmath,amssymb,
 aps,
 pre,
]{revtex4-2}

\usepackage{graphicx}
\usepackage{bm}
\usepackage[dvipsnames]{xcolor}
\usepackage{bbold}
\usepackage{mathtools}
\usepackage{xurl}
\usepackage{hyperref}

\begin{document}
\newcommand{\tr}{\text{Tr}}
\newcommand{\dpartial}{\dot{\partial}}
\newcommand{\bpartial}{\bar{\partial}}
\newcommand{\lagr}{\mathcal{L}}
\newcommand{\lagrgfm}{\mathcal{L}_{\text{GFM}}}
\newcommand{\lagrgf}{\mathcal{L}_{\text{GF}}}
\newcommand{\ham}{\mathcal{H}}
\newcommand{\hamgfm}{\mathcal{H}_{\text{GFM}}}
\newcommand{\hamgf}{\mathcal{H}_{\text{GF}}}
\newcommand{\attention}[1]{\textcolor{red}{#1}}
\newcommand{\ivan}[1]{\textcolor{red}{#1}}
\newcommand{\lit}{\textcolor{OliveGreen}{xxxADD CITATIONxxx }}
\newcommand{\morelit}{\textcolor{OliveGreen}{xxxADD MORE CITATIONSxxx }}
\newcommand{\secref}{\textcolor{purple}{xxxADD SECTIONxxx }}
\newcommand{\checkerr}{\textcolor{brown}{xxxCHECK EQUATIONSxxx }}
\newcommand{\leftbracket}{\bm{\{}\{}
\newcommand{\rightbracket}{\}\bm{\}}}
\newcommand{\PL}{\stackrel{\mathclap{\normalfont\mbox{pl}}}{=}}
\newcommand{\warning}[1]{\textcolor{Red}{#1}}

\title{Training time and depth as a \texorpdfstring{$1+1$}{1+1} field theory of learning}

\author{N. L. Holanda}
\email{linneuholanda@gmail.com}
\affiliation{Independent researcher, S\~{a}o Paulo, Brazil}

\author{I. E. Cunha}
\email{ivan.cunha@udesc.br}
\affiliation{Department of Biological Sciences, Santa Catarina State University (UDESC), Rua Coronel Fernandes Martins 270, Laguna, Santa Catarina 88790-000, Brazil}

\date{\today}

\begin{abstract}
We construct a field theory in $1+1$ dimensions for the training of deep residual networks, with the depth of the network as space and the training time as time. Its field equations, the master system, are derived as the continuum limit of gradient descent and of the heavy-ball method, for any width and any form of the blocks: at each depth the parameters obey the damped equation of motion of the optimizer, driven by the product of the error and the activation, while the activations and the errors obey the forward and adjoint equations along depth. The master system is the Hamilton system of a Bateman--Galley doubled action, in which the network is a set of constraints imposed by the errors as fields. In synchronous training, the forward and the backward pass are the secondary constraints of Dirac's algorithm, and backpropagation follows from the Dirac bracket. A finite speed of the signals along depth describes pipelined training: it turns the constraints into transport equations and makes the error the momentum of an activation, and the closed-time-path structure of the action selects the pairing in which the errors are causal. As solutions and limits of the master system, the theory recovers several results of the training theory of deep learning, among them backpropagation as an adjoint equation, the depth scaling of the learning rate, and the stability of delayed and pipelined training with weight prediction, and it gives a stability bound near a minimum that holds for any network: training is stable when the round trip of the signals is shorter than $\pi/2$ times the fastest relaxation time of the output. Numerical tests on deep tanh networks confirm the continuum limit, and a Dirac flow computed by automatic differentiation alone reproduces backpropagation in time and depth.
\end{abstract}

\maketitle

\section{Introduction}
\label{section_introduction}

A neural network is trained along two directions at once. Along its depth, a signal is passed forward through the layers and an error is passed back by backpropagation~\cite{rumelhart1986learning}; along the training time, the parameters move under the resulting gradient. Each direction has been described as the time of a dynamical system. Continuous-depth networks treat the layer index as time, so that the forward pass is an ordinary differential equation, backpropagation is its adjoint equation~\cite{chen2018neural}, and training is an optimal-control problem whose costate is the error~\cite{lecun1988theoretical, e2018mean}. Backpropagation has also been derived from a Bateman--Galley action in which the depth plays the role of time~\cite{scurria2026physical}. In Ref.~\cite{holanda2026backpropagation}, instead, the time is the training time: gradient-based optimizers are dissipative mechanical systems, given an action and a Hamiltonian by the Bateman--Galley doubling of the degrees of freedom~\cite{bateman1931dissipative, galley2013classical, galley2014principle}, the layers of the network are constraints on this system, and backpropagation follows from the Dirac bracket of the constraints~\cite{dirac1950generalized, henneaux1992quantization}.

Here we treat both directions as variables of a single field theory, with the depth $\tau \in [0, D]$ as space and the training time $t$ as time. The fields are the parameters $W(\tau, t)$, the activations $\mathbf{h}(\tau, t)$ and the errors $\bm\delta(\tau, t)$; the input enters at $\tau = 0$ and the cost at $\tau = D$. Earlier work with both variables includes continuum and mean-field limits of residual networks trained by gradient flow~\cite{lu2020mean, marion2023implicit}, which are coupled equations in depth and training time without an action, and an action over training time with the depth as a spatial coordinate, which does not contain backpropagation~\cite{lee2025synaptic}. Closest to the present work, Pehle and Slotine~\cite{pehle2026unlocked} relax the forward and backward passes as waves on a $(1+1)$-dimensional strip, with the depth as space and a finite propagation speed, and observe that the lock between the forward and the backward pass is an artifact of an infinite speed. Their construction starts from Pontryagin's principle and has no action, and it does not address the causal structure of the error transport or the stability of training.

The first question is which dynamics the field theory should describe. As in Ref.~\cite{holanda2026backpropagation}, where the master equation $\mu\ddot{\mathbf{x}} + \gamma\dot{\mathbf{x}} + \nabla L = 0$ is the continuum limit of the update rules, we derive the master equation of the $1+1$ theory as the continuum limit of training a residual network, with the layers becoming depth and the steps becoming training time. The result is local in depth and training time: at each depth the parameters obey the damped equation of motion of the optimizer, driven by the product of the error and the activation at the same depth, while the activations and the errors obey transport equations along depth. The master system plays for residual networks the role that a field equation plays for a physical system: it holds for any width and any form of the blocks, it is the Hamilton system of an action, and the results below are its solutions and limits. A single parameter, the speed $c$ at which signals cross the network in units of training time, controls the structure. In the limit $c\to\infty$, which we call the \emph{Coulomb limit}, the transport equations become constraints that hold at each training time, and the non-locality of backpropagation appears as the instantaneous limit of local propagation, as Coulomb's law is the instantaneous limit of retarded electromagnetic fields. A finite $c$ describes pipelined, asynchronous and physical learning, in which the parameters change while the signals cross the network.

Our construction follows Ref.~\cite{holanda2026backpropagation}, and our results hold for deep residual networks of any width and with any form of the blocks. (i) The continuum limit of gradient descent and of the heavy-ball method is the master system \eqref{master_weights}--\eqref{master_adjoint}, that of Nesterov's method is the same system with the force evaluated at look-ahead fields, and the limit exists only if the per-parameter learning rate grows in proportion to the number of layers. The Bateman--Galley doubled action \eqref{doubled_action_cinf} of the master system carries the damping of the optimizer on the parameters and the network as constraints imposed by the errors, and we give its doubled Legendre transform and Hamiltonian; Nesterov's method and the weight prediction of pipelined training enter through look-ahead fields with a look-ahead time, as in Ref.~\cite{holanda2026backpropagation}, and for gradient descent through the second-class constraints of gradient flow; at the time of the momentum, the look-ahead fields of the heavy ball carry the Dirac bracket of gradient flow (Sec.~\ref{section_damped_fields}). (ii) The forward pass and the backward pass of each copy are the secondary constraints of Dirac's algorithm; the Dirac bracket reduces the theory to the phase space of the parameters, with the activations as functionals of the parameters and a retarded kernel along depth, the bracket of the output with the momentum of a parameter is the Jacobian of the network, and the physical limit of the Dirac evolution is the damped master system (Sec.~\ref{section_depth_constraint}). (iii) For pipelined training, $c$ is the inverse of the per-parameter learning rate, the continuum limit is a system of transport equations, and the errors travel back along depth forward in training time, in contrast to the exact gradient of the loss accumulated over training, which requires them to run backward in training time, a problem known from learning in physical time~\cite{ellenberger2026backpropagation}. In the doubled action, a finite $c$ makes the error the momentum of an activation: pairing it with the activation of its own copy gives the exact gradient, and pairing it with the activation of the other copy, as in the damping term of Ref.~\cite{holanda2026backpropagation}, gives the causal algorithm. The two pairings have the same Hamiltonian and differ only in the bracket, and the closed-time-path structure of the action, an initial-value problem in training time with the boundary condition at the output, admits only causal pairings, of which the crossed one is the only one with equal speeds of the activations and the errors (Sec.~\ref{section_finite_c}). (iv) Near an interpolating minimum, the delays turn gradient flow into a neural-tangent-kernel equation in which each depth contributes with the delay of its round trip. The delay acts as an inertia set by the depth profile of the kernel, the implicit momentum of asynchronous training~\cite{mitliagkas2016asynchrony}, and training is stable if the round trip of the signals is shorter than $\pi/2$ times the fastest relaxation time of the output; beyond a threshold that depends on the profile, the minimum loses stability through a Hopf bifurcation (Sec.~\ref{section_stability}). In gradient descent, a look-ahead of part of the round trip, a weight prediction, raises the threshold, while a prediction over the full delay overcompensates the delay, and the velocity estimate of the pipeline adds an alternating mode that limits large look-aheads. With momentum, a look-ahead over the time of the momentum turns the pipelined heavy ball exactly into delayed gradient descent with an exponential moving average of the parameters, for the full nonlinear network; Nesterov's look-ahead is too short for this, and beyond the momentum time the best look-ahead grows with the momentum (Sec.~\ref{section_lookahead}). (v) On deep tanh networks, the parameters of gradient descent, the heavy ball, Nesterov's method and the pipeline converge to the field equations at first order in the layer spacing, a Dirac flow computed by automatic differentiation alone, without a coded backward pass, reproduces backpropagation in time and depth, and the threshold of stability agrees with the delayed kernel (Sec.~\ref{section_numerical_tests}).

Many of these results are known in optimization and machine learning, where they were found one at a time and by different methods. Here they follow from a single action principle, as solutions and limits of the master system, and agree exactly with the published formulas (Table~\ref{table_known_results}). We regard this agreement as the evidence for the field theory. The contribution of the present work is the physical structure: the master system as the field equations of training, its doubled action, Hamiltonian and Dirac brackets over depth and training time, the pairings that separate the exact gradient from the causal algorithm, and the transport of the errors at a finite speed. The optimization results in Table~\ref{table_known_results} without a reference we have not found stated elsewhere, and we present them, like the others, as consequences of the theory.

\begin{table*}[tb]
\caption{\label{table_known_results}Results of optimization and machine learning that follow from the field theory, with the part of the theory that gives them. References are given where the result has been published.}
\begin{ruledtabular}
\begin{tabular}{p{0.46\textwidth}p{0.14\textwidth}p{0.3\textwidth}}
\raggedright Result & \raggedright Refs. & \raggedright In the field theory \tabularnewline
\hline
\raggedright Backpropagation as the adjoint (costate) equation of the forward pass & \raggedright \cite{lecun1988theoretical, chen2018neural} & \raggedright Dirac bracket of the depth constraint, Coulomb limit, Sec.~\ref{section_depth_constraint} \tabularnewline
\raggedright Gradient descent, the heavy ball and Nesterov's method as damped dynamics & \raggedright \cite{polyak1964some, nesterov1983method, holanda2026backpropagation} & \raggedright master system and look-ahead fields, Sec.~\ref{section_damped_fields} \tabularnewline
\raggedright Weight prediction in pipelined training & \raggedright \cite{chen2018efficient, kosson2020pipelined} & \raggedright look-ahead fields with a look-ahead time, Sec.~\ref{section_nesterov} \tabularnewline
\raggedright Residual networks with branches scaled by $1/L$ need a learning rate that grows like $L$ & \raggedright \cite{marion2023implicit} & \raggedright depth-scaled rate, Eq.~\eqref{depth_scaled_rates} \tabularnewline
\raggedright Causal learning sends the errors forward in time; the exact gradient needs them backward & \raggedright \cite{ellenberger2026backpropagation} & \raggedright crossed and direct pairings, Sec.~\ref{section_finite_c} \tabularnewline
\raggedright Linearized training as gradient flow with the neural tangent kernel & \raggedright \cite{jacot2018neural} & \raggedright Coulomb limit of Eq.~\eqref{delayed_ntk} \tabularnewline
\raggedright Stale gradients act as momentum & \raggedright \cite{mitliagkas2016asynchrony} & \raggedright inertia $(2/c^2)K_2H$, Eq.~\eqref{delay_inertia_general} \tabularnewline
\raggedright Delays destabilize gradient flow through a Hopf bifurcation & \raggedright \cite{janson2021delay} & \raggedright normal modes of the delayed kernel, Sec.~\ref{section_stability} \tabularnewline
\raggedright Training of any network is stable if the round trip is shorter than $\pi/2$ times the fastest relaxation time of the output & \raggedright --- & \raggedright positivity of the delayed kernel, Eq.~\eqref{stability_bound} \tabularnewline
\raggedright The delays and the threshold depend on the depth at which the kernel is accumulated & \raggedright --- & \raggedright depth profile of the kernel, Sec.~\ref{section_stability} \tabularnewline
\raggedright Threshold of delayed gradient descent, $g = 2\sin[\pi/(4\tau + 2)] \approx \pi/(2\tau)$ & \raggedright \cite{giladi2019stability} & \raggedright kernel at a single depth, Sec.~\ref{section_stability} \tabularnewline
\raggedright Momentum shrinks the stable range under a delay & \raggedright \cite{giladi2019stability} & \raggedright heavy ball with delay, Sec.~\ref{section_stability} \tabularnewline
\raggedright Characteristic polynomial of linear weight prediction without momentum & \raggedright \cite{kosson2020pipelined} & \raggedright look-ahead at a single depth, Sec.~\ref{section_lookahead} \tabularnewline
\raggedright In gradient descent, a partial prediction raises the threshold and a prediction over the full delay overcompensates it & \raggedright --- & \raggedright look-ahead kernel, Eq.~\eqref{lookahead_kernel} \tabularnewline
\raggedright Momentum with the gradient at an extrapolated point is gradient descent with weight averaging & \raggedright \cite{defazio2024road, morwani2025connections} & \raggedright look-ahead fields at $\varepsilon = \mu/\gamma$, with the bracket of gradient flow, Secs.~\ref{section_nesterov} and \ref{section_lookahead} \tabularnewline
\raggedright With a delay, the same look-ahead gives delayed gradient descent with weight averaging, for a single delay and a quadratic loss (from the equivalence of weight prediction and spike compensation) & \raggedright \cite{kosson2020pipelined} & \raggedright look-ahead over the momentum time, Sec.~\ref{section_lookahead} \tabularnewline
\raggedright The same holds exactly for the nonlinear pipeline with delays spread over depth &  \raggedright --- & \raggedright look-ahead over the momentum time, Sec.~\ref{section_lookahead} \tabularnewline
\raggedright With momentum, weight prediction enlarges the stable region, horizons beyond the delay converge faster, and Nesterov's look-ahead helps little & \raggedright \cite{kosson2020pipelined} & \raggedright look-ahead with momentum, Sec.~\ref{section_lookahead} \tabularnewline
\end{tabular}
\end{ruledtabular}
\end{table*}

The paper is organized as follows. Section~\ref{section_damped_fields} derives the master system and gives its doubled action and Hamiltonian, Sec.~\ref{section_depth_constraint} derives backpropagation from the Dirac bracket of the depth constraint, Sec.~\ref{section_finite_c} treats a finite propagation speed, its brackets and the stability of training, and Sec.~\ref{section_numerical_tests} reports the numerical tests. We conclude in Sec.~\ref{section_conclusion}. Appendix~\ref{appendix_gradient_flow} treats gradient flow and Appendix~\ref{appendix_brackets} lists the complete brackets. The numbers and figures of Sec.~\ref{section_numerical_tests} are produced by code that accompanies the paper~\cite{holanda2026twotime}, and every equation has been checked symbolically.

\section{Training as damped field dynamics over depth}
\label{section_damped_fields}

\subsection{Continuum limit of a residual network}
\label{section_master_equation}
Consider a residual network with $N$ layers,
\begin{equation}
\label{residual_layers}
\mathbf{h}_k = \mathbf{h}_{k-1} + \Delta\tau\,\mathbf{f}(W_k, \mathbf{h}_{k-1}),
\end{equation}
for $k = 1, \dots, N$, with $\mathbf{h}_0 = \mathbf{x}$, where $\Delta\tau = D/N$, $W_k$ are the parameters of layer $k$, and the loss is the cost of the output, $L = \ell(\mathbf{h}_N)$, averaged over the samples, whose index we suppress. Layer $k$ sits at the depth $\tau_k = k\,\Delta\tau$, and as $N\to\infty$ the network becomes the continuous-depth network $\partial_\tau\mathbf{h} = \mathbf{f}(W(\tau), \mathbf{h})$~\cite{chen2018neural}. Backpropagation gives the gradient
\begin{equation}
\label{discrete_gradient}
\begin{split}
&\frac{\partial L}{\partial W_k} = \Delta\tau\Big(\frac{\partial\mathbf{f}}{\partial W}\Big)_k^T\hat{\bm\delta}_k,\\
&\hat{\bm\delta}_N = \frac{\partial\ell}{\partial\mathbf{h}_N},\qquad \hat{\bm\delta}_k = \big(\mathbb{I} + \Delta\tau J_{k+1}\big)^T\hat{\bm\delta}_{k+1},
\end{split}
\end{equation}
where $J_k = \partial\mathbf{f}/\partial\mathbf{h}$ and $(\partial\mathbf{f}/\partial W)_k$ are evaluated at $(W_k, \mathbf{h}_{k-1})$. The errors $\hat{\bm\delta}_k$ are normalized so that they have a limit as $\Delta\tau\to0$: the backward recursion is the explicit Euler discretization of the adjoint equation $-\partial_\tau\bm\delta = J^T\bm\delta$. The gradient with respect to the parameters of layer $k$ is therefore $\Delta\tau$ times a density that depends only on the activation and the error at the depth of the layer.

The optimizers are those of Ref.~\cite{holanda2026backpropagation}: gradient descent with learning rate $\lambda$, $W_k[n+1] = W_k[n] - \lambda\,\partial L/\partial W_k$, the heavy-ball method~\cite{polyak1964some}, $W_k[n+1] = W_k[n] + \beta(W_k[n] - W_k[n-1]) - \alpha\,\partial L/\partial W_k$, and Nesterov's method~\cite{nesterov1983method}, which evaluates the gradient at the look-ahead point $W_k[n] + \beta(W_k[n] - W_k[n-1])$ (Sec.~\ref{section_nesterov}). A step advances the training time by $h$. With
\begin{equation}
\label{depth_scaled_rates}
\lambda = \frac{h}{\gamma\,\Delta\tau},\quad \alpha = \frac{h^2}{\mu\,\Delta\tau},\quad \beta = 1 - \frac{h\gamma}{\mu},
\end{equation}
both update rules are consistent discretizations, as $h\to0$ and $\Delta\tau\to0$, of the \emph{master system}
\begin{subequations}
\label{master_system}
\begin{align}
\label{master_weights}
&\mu\,\partial_t^2W + \gamma\,\partial_tW = -\Big(\frac{\partial\mathbf{f}}{\partial W}\Big)^T\bm\delta ,\\
\label{master_forward}
&\partial_\tau\mathbf{h} = \mathbf{f}(W, \mathbf{h}),\qquad \mathbf{h}(0, t) = \mathbf{x},\\
\label{master_adjoint}
&-\partial_\tau\bm\delta = J^T\bm\delta,\qquad \bm\delta(D, t) = \frac{\partial\ell}{\partial\mathbf{h}(D, t)},
\end{align}
\end{subequations}
with $\mu = 0$ for gradient descent. All fields depend on $(\tau, t)$, and every equation is local: the force on the parameters at depth $\tau$ is the product of the error and the activation at the same depth, $(\sigma'\odot\bm\delta)\,\mathbf{h}^T$ for a layer $\mathbf{f} = \sigma(W\mathbf{h})$. For the parameters alone, equation \eqref{master_weights} reads $\mu\,\partial_t^2W + \gamma\,\partial_tW = -\delta L/\delta W(\tau)$, an integro-differential equation in depth, since the functional gradient depends on the parameters at all depths. The non-locality of backpropagation is carried entirely by equations \eqref{master_forward} and \eqref{master_adjoint}, which hold at each training time and transmit a change of the parameters at one depth instantly to all others. In Sec.~\ref{section_finite_c} they become transport equations with a finite speed $c$, and the system \eqref{master_system} is their Coulomb limit $c\to\infty$. Nothing in the system depends on the width of the network or on the form of the blocks.

The scaling \eqref{depth_scaled_rates} is necessary. At a fixed per-parameter learning rate $\lambda$, one step of gradient descent advances the training time by $h = \gamma\lambda\,\Delta\tau$, which vanishes as $N\to\infty$: the dynamics freezes. A residual network with branches scaled by $1/N$ therefore has a continuum limit in training time only if its learning rate grows in proportion to its depth. This is the scaling under which residual networks approach neural ordinary differential equations during training: in terms of the gradient of the loss, the rate \eqref{depth_scaled_rates} gives $\partial_tW_k = -(N/\gamma D)\,\partial L/\partial W_k$, which for $D = \gamma = 1$ is the gradient flow with the factor $L = N$ of Ref.~\cite{marion2023implicit}, introduced there to prevent vanishing gradients as $L\to\infty$; for the gradient flow, Ref.~\cite{marion2023implicit} proves that the parameters converge to a limit in depth and training time. The depth-$\mu$P parametrization~\cite{yang2023tensor, bordelon2023depthwise} scales the branches by $1/\sqrt N$ instead, with learning rates adapted to that choice. The parameters of deep networks converge to the master system at first order in $\Delta\tau$, for gradient descent and for the heavy ball (Sec.~\ref{section_numerical_tests}).

\subsection{Bateman--Galley doubling}
Equation \eqref{master_weights} is dissipative, and as in Ref.~\cite{holanda2026backpropagation} we double every field, $(W_i, \mathbf{h}_i, \bm\delta_i)$ with $i = 1, 2$~\cite{bateman1931dissipative, galley2013classical, galley2014principle}. We use the covariant notation of Ref.~\cite{holanda2026backpropagation}: $\mathbf{h}^a = (\mathbf{h}_1, \mathbf{h}_2)$, $\bm\delta_a = (\bm\delta_1, -\bm\delta_2)$, the metric $\eta = \mathrm{diag}(1, -1)$, the exchange matrix $X$ and the conjugate $\mathbf{h}^{\dagger a} = X^a{}_b\mathbf{h}^b = (\mathbf{h}_2, \mathbf{h}_1)$. The parameters at each depth carry the doubled Lagrangian of Ref.~\cite{holanda2026backpropagation} (for constant $\mu$ and $\gamma$), the forward pass of each copy is a constraint imposed by its error as a multiplier field, and the cost enters at the output boundary through the antisymmetric potential $V = \ell(\mathbf{h}_1(D)) - \ell(\mathbf{h}_2(D))$. The action of the Coulomb limit is $\int\Lambda_\infty\,dt$ with
\begin{equation}
\label{doubled_action_cinf}
\begin{split}
\Lambda_\infty = \int_0^D\Big[&\frac\mu2\big(\dot W_1^2 - \dot W_2^2\big) + \frac\gamma2\big(\dot W_1W_2 - \dot W_2W_1\big)\\
&+ \bm\delta_a\cdot\big(\partial_\tau\mathbf{h}^a - \mathbf{f}(W^a, \mathbf{h}^a)\big)\Big]d\tau - V ,
\end{split}
\end{equation}
where the dot is $\partial_t$. Its Euler--Lagrange equations are exact for the doubled system. Varying $\bm\delta_i$ gives the depth constraint \eqref{master_forward} of copy $i$, varying $\mathbf{h}_i$ gives the adjoint equation \eqref{master_adjoint} of copy $i$, and the boundary term of this variation at the output, $\pm\big(\bm\delta_i(D) - \partial\ell/\partial\mathbf{h}_i(D)\big)\cdot\delta\mathbf{h}_i(D)$, vanishes exactly for the boundary condition of the adjoint equation, while the input $\mathbf{h}(0) = \mathbf{x}$ is fixed. Varying $W_i$ gives the doubled equation of motion of the parameters. On the physical limit $W_1 = W_2$, $\mathbf{h}_1 = \mathbf{h}_2$, $\bm\delta_1 = \bm\delta_2$, which is invariant, both copies obey the damped master system \eqref{master_system}. The constraint sector, $\bm\delta_a\cdot(\partial_\tau\mathbf{h}^a - \mathbf{f}^a)$, is the same for every optimizer: the optimizer enters only through the Lagrangian of the parameters, and the network only through the constraints.

\emph{What the action determines.} The Euler--Lagrange equations of the action \eqref{doubled_action_cinf} are the field equations of both copies, and the condition at the output, $\bm\delta(D, t) = \partial\ell/\partial\mathbf{h}(D, t)$, is its natural boundary condition: the cost enters only through the potential $V$ at $\tau = D$, and the boundary term of the variation vanishes for it. Imposed as data are the input $\mathbf{h}(0, t) = \mathbf{x}$, which is held fixed in the variation, the initial parameters $W(\tau, 0)$ and, for $\mu \neq 0$, their velocities, and, at finite $c$, the initial activations and errors in the pipeline (Sec.~\ref{section_finite_c}). The action is constructed to reproduce the master system, which we derive from the update rules: its Lagrangian of the parameters is that of Ref.~\cite{holanda2026backpropagation}, its constraint sector is fixed by the network, and at finite $c$ the pairing of the errors with the activations is fixed by causality, which the closed-time-path structure of the action requires, and by the equal speeds of the activations and the errors (Sec.~\ref{section_canonical_structure}). The look-ahead of Nesterov's method and of weight prediction follows from the action through look-ahead fields (Sec.~\ref{section_nesterov}). The closed-time-path conditions at the initial and final training times~\cite{galley2013classical} hold for every pairing and impose no final condition on the physical error.

\subsection{Hamiltonian and brackets}
The momenta conjugate to the parameters are those of Ref.~\cite{holanda2026backpropagation}, $P_1 = \mu\dot W_1 + \frac\gamma2W_2$ and $P_2 = -\mu\dot W_2 - \frac\gamma2W_1$. The action contains no time derivatives of the activations and the errors, so that their momenta vanish,
\begin{equation}
\label{primary_constraints_cinf}
\bm\pi_i = \frac{\partial\Lambda_\infty}{\partial\dot{\mathbf{h}}_i} \approx 0,\qquad \bm\rho_i = \frac{\partial\Lambda_\infty}{\partial\dot{\bm\delta}_i} \approx 0 ,
\end{equation}
which are primary constraints. We use plain canonical momenta, with the Poisson brackets $\{W_i(\tau, t), P_j(\tau', t)\} = \{h_i(\tau, t), \pi_j(\tau', t)\} = \{\delta_i(\tau, t), \rho_j(\tau', t)\} = \delta_{ij}\,\delta(\tau - \tau')$ between fields at equal training time, and all others zero; this is the bracket $\Gamma = \Gamma_1 - \Gamma_2$ of Ref.~\cite{holanda2026backpropagation}, written in its covariant momenta. The doubled Legendre transform is
\begin{equation}
\label{doubled_legendre}
\begin{split}
\mathcal{H}_\infty &= \int_0^D\sum_{i=1}^2\big(P_i\dot W_i + \bm\pi_i\cdot\dot{\mathbf{h}}_i + \bm\rho_i\cdot\dot{\bm\delta}_i\big)\,d\tau - \Lambda_\infty\\
&= \int_0^D\Big[\frac\mu2\big(\dot W_1^2 - \dot W_2^2\big) - \bm\delta_a\cdot\big(\partial_\tau\mathbf{h}^a - \mathbf{f}^a\big)\Big]d\tau + V,
\end{split}
\end{equation}
on the primary constraints, where the terms of $P_i\dot W_i$ linear in the velocities cancel the damping term of $\Lambda_\infty$. With $\dot W_1 = (P_1 - \frac\gamma2W_2)/\mu$ and $\dot W_2 = -(P_2 + \frac\gamma2W_1)/\mu$, the canonical Hamiltonian is
\begin{equation}
\label{doubled_hamiltonian}
\begin{split}
\mathcal{H}_\infty = \int_0^D\Big[&\frac{(P_1 - \frac\gamma2W_2)^2 - (P_2 + \frac\gamma2W_1)^2}{2\mu}\\
&- \bm\delta_a\cdot\big(\partial_\tau\mathbf{h}^a - \mathbf{f}(W^a, \mathbf{h}^a)\big)\Big]d\tau + V ,
\end{split}
\end{equation}
the Hamiltonian of the parameters of Ref.~\cite{holanda2026backpropagation} at each depth, plus the constraint terms of the two copies and the potential at the output. Unlike in Ref.~\cite{holanda2026backpropagation}, where the errors appear as the Lagrange multipliers of the layer constraints, here they are fields of phase space, and the forward and backward passes follow from Dirac's algorithm, which we apply in the next section. For gradient descent, $\mu = 0$, the Legendre transform of the parameters is singular as well (Appendix~\ref{appendix_gradient_flow}).

\subsection{Look-ahead fields}
\label{section_nesterov}
Nesterov's method evaluates the network at the look-ahead parameters $W + \beta(W[n] - W[n-1]) \approx W + h\,\partial_tW$, and pipelined training predicts the parameters over longer times (Sec.~\ref{section_lookahead}). As in Ref.~\cite{holanda2026backpropagation}, we describe both by look-ahead parameter fields $Y_i(\tau, t)$ with momenta $\bm\varpi_i$, at which the network of each copy, forward and adjoint, is evaluated: the constraint terms of the Hamiltonian \eqref{doubled_hamiltonian} become $-\bm\delta_a\cdot(\partial_\tau\mathbf{h}^a - \mathbf{f}(Y^a, \mathbf{h}^a))$, and the look-ahead fields are fixed by the constraints
\begin{equation}
\label{lookahead_constraints}
\psi_i = Y_i - W_i - \varepsilon\,\bar v \approx 0,\qquad \bm\varpi_i \approx 0,
\end{equation}
where $\varepsilon(\tau)$ is a look-ahead time at each depth, $\varepsilon = h$ for Nesterov's method, and $\bar v = (v_1 + v_2)/2$ is the mean of the velocities $v_i = \partial\mathcal{H}_\infty/\partial P_i$ of the two copies. The constraints are local in depth and do not change the network, and the look-ahead fields of the two copies do not commute, $\{Y_1(\tau, t), Y_2(\tau', t)\}_D = (\varepsilon/\mu)\,\delta(\tau - \tau')$, as in Ref.~\cite{holanda2026backpropagation} for $\varepsilon = h$. In the physical limit, the Dirac evolution gives
\begin{equation}
\label{nesterov_field}
\mu\,\partial_t^2W + \gamma\,\partial_tW = -\Big(\frac{\partial\mathbf{f}}{\partial W}\Big)^T\Big|_Y\bm\delta[Y],\qquad Y = W + \varepsilon\,\partial_tW,
\end{equation}
in which the force is the gradient density at the look-ahead fields, while the activations follow the look-ahead fields through the forward pass. For Nesterov's method, $h$ is the step of the optimizer, which stays finite in the continuum limit in depth.

At the time of the momentum, $\varepsilon = \mu/\gamma$, the look-ahead fields have the bracket $\{Y_1(\tau, t), Y_2(\tau', t)\}_D = \gamma^{-1}\delta(\tau - \tau')$ of the parameters of gradient flow (Appendix~\ref{appendix_gradient_flow}), and their Dirac evolution is gradient flow, $\gamma\,\partial_tY = -(\partial\mathbf{f}/\partial W)^T|_Y\bm\delta[Y]$: the heavy ball contains a gradient-flow theory in its look-ahead fields, which the parameters follow through $(\mu/\gamma)\,\partial_tW + W = Y$ (Sec.~\ref{section_lookahead}). For gradient descent, $\mu = 0$, the look-ahead fields are not a limit of these, since $\{Y_1, Y_2\}_D = \varepsilon/\mu$ diverges. We construct them in the phase space of gradient flow, with its second-class constraints (Appendix~\ref{appendix_gradient_flow}), and with the velocity of the gradient-flow Dirac evolution, $v_i = -\gamma^{-1}(\partial\mathbf{f}/\partial W)^T|_{Y_i}\bm\delta_i$, in the constraints \eqref{lookahead_constraints}. The constraints remain second class wherever $Y = W + \varepsilon\,\partial_tW$ can be solved for $Y$, the parameters keep the bracket of gradient flow, and the physical limit is $\gamma\,\partial_tW = -(\partial\mathbf{f}/\partial W)^T|_Y\bm\delta[Y]$ with $Y = W + \varepsilon\,\partial_tW$. The brackets of the look-ahead fields are then not local in depth, since the velocity contains the errors: for a constant $\varepsilon$, $\{Y_1, Y_2\}_D$ is the kernel $\gamma^{-1}[1 + (\varepsilon/\gamma)\,\delta^2L/\delta W\delta W]^{-1}$, the bracket of gradient flow carried by the map from $W$ to $Y$. At finite $c$ (Sec.~\ref{section_finite_c}), the look-ahead constraints of the heavy ball commute with those of the activations and errors, the bracket $\{Y_1, Y_2\}_D = (\varepsilon/\mu)\,\delta(\tau - \tau')$ is unchanged, and for both optimizers the physical limit is the same equation of the parameters with the transport driven by the look-ahead fields.

\section{Backpropagation as Dirac evolution}
\label{section_depth_constraint}

\subsection{Layers as constraints}
Discretized as in equation \eqref{residual_layers}, the constraint term of copy $i$ is $\bm\phi_{i,k} = (\mathbf{h}_{i,k} - \mathbf{h}_{i,k-1})/\Delta\tau - \mathbf{f}(W_{i,k}, \mathbf{h}_{i,k-1}) \approx 0$, which is the layer constraint of Ref.~\cite{holanda2026backpropagation} for a residual network divided by $\Delta\tau$; the division does not change the Dirac bracket. Dirac's algorithm~\cite{dirac1950generalized, henneaux1992quantization, gitman1990quantization} requires the primary constraints \eqref{primary_constraints_cinf} to be preserved in training time. Since $\mathcal{H}_\infty$ depends on the errors only through the constraint terms, preserving $\bm\rho_i \approx 0$ gives the depth constraint of copy $i$, $\partial_\tau\mathbf{h}_i - \mathbf{f}(W_i, \mathbf{h}_i) \approx 0$, equation \eqref{master_forward}. Preserving $\bm\pi_i \approx 0$ gives, after an integration by parts in depth, the adjoint equation $-\partial_\tau\bm\delta_i - J_i^T\bm\delta_i \approx 0$ with the boundary condition $\bm\delta_i(D) = \partial\ell/\partial\mathbf{h}_i(D)$, equation \eqref{master_adjoint}; discretized, it is the backward recursion \eqref{discrete_gradient}. These secondary constraints contain no momenta, and the algorithm stops: the forward pass and the backward pass of each copy are the secondary constraints of the doubled theory. The constraints $\bm\pi_i$, $\bm\rho_i$, the depth constraints and the adjoint equations are second class; their matrix of brackets is triangular in depth, with the identity on its diagonal blocks, so that its determinant is $1$ for any width and any block. Since the constraints of the two copies have vanishing brackets with each other, the constraint matrix is block diagonal over the copies.

\subsection{The Dirac bracket}
On the constraint surface, the activations and the errors of copy $i$ are functionals of its parameters, $\mathbf{h}_i = \mathbf{h}[W_i]$ and $\bm\delta_i = \bm\delta[W_i]$, the constraint terms vanish, and the Hamiltonian \eqref{doubled_hamiltonian} becomes
\begin{equation}
\label{reduced_hamiltonian}
\begin{split}
\mathcal{H}_\infty = &\int_0^D\frac{(P_1 - \frac\gamma2W_2)^2 - (P_2 + \frac\gamma2W_1)^2}{2\mu}\,d\tau\\
&+ L[W_1] - L[W_2],
\end{split}
\end{equation}
with the loss $L[W] = \ell(\mathbf{h}(D; W))$: the doubled Hamiltonian of Ref.~\cite{holanda2026backpropagation}, for parameters that are fields over depth. The Dirac bracket is the canonical bracket of this reduced phase space. For functionals $F$ and $G$ of the parameters, activations and errors, at equal training time,
\begin{equation}
\label{dirac_reduced_cinf}
\begin{split}
&\big\{F, P_i(\tau', t)\big\}_D = \frac{\delta F}{\delta W_i(\tau')},\\
&\{F, G\}_D = 0,\qquad \big\{P_i(\tau, t), P_j(\tau', t)\big\}_D = 0 .
\end{split}
\end{equation}
For the activations this is the continuum form of the Jacobian bracket of Ref.~\cite{holanda2026backpropagation},
\begin{equation}
\label{retarded_kernel}
\big\{\mathbf{h}_i(\tau, t), P_j(\tau', t)\big\}_D = \delta_{ij}\,\vartheta(\tau - \tau')\,\frac{\delta\mathbf{h}(\tau)}{\delta W(\tau')}\bigg|_{W_i},
\end{equation}
which vanishes for $\tau' > \tau$, since an activation depends only on the parameters before it; here $\vartheta$ is the step function. An error, in contrast, depends on the parameters both before its depth, through the activations, and after it, and its bracket with $P_i(\tau')$ is in general nonzero on both sides of $\tau$. At the output, the bracket of every component of $\mathbf{h}_i(D)$ with the momentum of every parameter of its copy is the Jacobian of the network, $\{\mathbf{h}_i(D), P_{i,\theta}\}_D = \partial\mathbf{h}(D)/\partial\theta$. The Dirac evolution of the momenta is therefore $\dot P_1 = -\frac\gamma2\dot W_2 - \delta L/\delta W_1$ and its mirror for copy $2$, and with $P_1 = \mu\dot W_1 + \frac\gamma2W_2$ the physical limit is
\begin{equation}
\label{dirac_master}
\mu\,\partial_t^2W + \gamma\,\partial_tW = -\frac{\delta L}{\delta W(\tau)} = -\Big(\frac{\partial\mathbf{f}}{\partial W}\Big)^T\bm\delta ,
\end{equation}
with the activations and errors given by the forward and backward passes: the Dirac bracket of the depth constraint backpropagates through a deep nonlinear network, as the Dirac bracket of the layer constraints does in Ref.~\cite{holanda2026backpropagation}, and the damped master system \eqref{master_system} is the physical limit of the Dirac evolution. The derivation uses only the structure of the constraints and holds for any width, block, activation function and cost. Section~\ref{section_dirac_test} tests it numerically.

\subsection{Mechanical interpretation}
In the Coulomb limit, the activations and the errors have no dynamics of their own. They are fixed at each training time by the parameters, as the Coulomb field of electrostatics is fixed by the charges, and a change of the parameters at one depth reaches every other depth instantly. The bracket \eqref{retarded_kernel} refines the analogy: the Coulomb bracket of a gauge field is non-local in every direction, while the depth constraint gives a kernel that is retarded along depth and instantaneous in training time. The forward pass propagates the kernel from the input to the output, and the backward pass, the preservation of $\bm\pi_i \approx 0$, transports the cost from the output back to every depth. Since the constraint sector does not involve the parameter Lagrangian, the forward and backward passes are the same for every optimizer, and the Dirac bracket reduces any damped motion of the parameters to its motion on the loss landscape $L[W]$. For gradient flow, $\mu = 0$, the constraints of the parameters do not commute with those of the network, and the bracket is built in stages (Appendix~\ref{appendix_gradient_flow}). For Nesterov's method, the look-ahead constraints \eqref{lookahead_constraints} enter between the parameters and the network: the constraint algebra keeps the structure above, with the depth constraint and the adjoint equation evaluated at the look-ahead fields, all constraints remain second class, and the Dirac evolution of the momenta is the force at the look-ahead fields, equation \eqref{nesterov_field}.

\section{Finite signal speed}
\label{section_finite_c}

\subsection{Pipelined training}
In the Coulomb limit, a change of the parameters at one depth reaches every other depth instantly in training time. This is the case of ordinary, synchronous training, in which the forward and backward passes are completed before the parameters move. In pipeline-parallel and asynchronous training, and in physical learning machines, the parameters change while the signals cross the network. We define $c$ as the speed of the forward signal along depth, in units of depth per unit of training time: a signal that crosses one layer, of depth $\Delta\tau$, during one step, of training time $h$, has $c = \Delta\tau/h$.

The simplest model is pipelined gradient descent with one layer per step and without weight stashing~\cite{harlap2018pipedream}; the staleness it causes, and corrections for it, are studied in Refs.~\cite{yang2019pipemare, kosson2020pipelined}. At every step, all layers act at once, with the values of the previous step on the right-hand sides:
\begin{equation}
\label{pipeline_update}
\begin{split}
\mathbf{h}_k &\leftarrow \mathbf{h}_{k-1} + \Delta\tau\,\mathbf{f}(W_k, \mathbf{h}_{k-1}),\\
\bm\delta_k &\leftarrow \big(\mathbb{I} + \Delta\tau J_{k+1}\big)^T\bm\delta_{k+1},\\
W_k &\leftarrow W_k - \frac{h}{\gamma}\Big(\frac{\partial\mathbf{f}}{\partial W}\Big)_k^T\bm\delta_k,
\end{split}
\end{equation}
with $\mathbf{h}_0 = \mathbf{x}$, $\bm\delta_{N+1} = \partial\ell/\partial\mathbf{h}_N$ and $J_{N+1} = 0$. Each activation moves forward by one layer, each error moves back by one layer, and each layer updates its parameters with its current activation and error. The per-parameter learning rate of the update is $\lambda = h/(\gamma\Delta\tau)$, as in equation \eqref{depth_scaled_rates}, so that
\begin{equation}
\label{c_learning_rate}
c = \frac{\Delta\tau}{h} = \frac{1}{\gamma\lambda},
\end{equation}
and $c = m/(\gamma\lambda)$ for a pipeline that moves $m$ layers per step. The learning rate sets the unit of training time, and in a pipeline it fixes the speed of the signals in that unit: synchronous training is the limit of small learning rates. The continuum limit at fixed $c$, with $\Delta\tau\to0$ and $h = \Delta\tau/c\to0$ together, is the master system at finite $c$,
\begin{subequations}
\label{finite_c_system}
\begin{align}
\label{finite_c_weights}
&\gamma\,\partial_tW = -\Big(\frac{\partial\mathbf{f}}{\partial W}\Big)^T\bm\delta ,\\
\label{finite_c_forward}
&(\partial_t + c\,\partial_\tau)\,\mathbf{h} = c\,\mathbf{f}(W, \mathbf{h}),\qquad \mathbf{h}(0, t) = \mathbf{x},\\
\label{finite_c_errors}
&(\partial_t - c\,\partial_\tau)\,\bm\delta = c\,J^T\bm\delta,\qquad \bm\delta(D, t) = \frac{\partial\ell}{\partial\mathbf{h}(D, t)} .
\end{align}
\end{subequations}
For the heavy ball, with the rates \eqref{depth_scaled_rates} at $h = \Delta\tau/c$, the weight equation \eqref{finite_c_weights} becomes $\mu\,\partial_t^2W + \gamma\,\partial_tW = -(\partial\mathbf{f}/\partial W)^T\bm\delta$, with the same transport equations (Sec.~\ref{section_numerical_tests}). Dividing the transport equations by $c$ shows that they reduce to the constraints \eqref{master_forward} and \eqref{master_adjoint} in the Coulomb limit, with corrections of order $1/c$. All three equations are local and first order. The activations move forward along the characteristics $\tau - ct = $ const and the errors move backward along $\tau + ct = $ const, both forward in training time: the error transport of the algorithm is causal. The exact gradient of the loss accumulated over training, $\sum_n\ell(\mathbf{h}_N[n])$, obeys instead the advanced recursion obtained from equation \eqref{pipeline_update} by reversing the step index of the errors, whose continuum limit is $-(\partial_t + c\,\partial_\tau)\tilde{\bm\delta} = c\,J^T\tilde{\bm\delta}$: backpropagation through training time, which runs backward in $t$. With frozen parameters, the pipeline forgets its initial state after $2N$ steps, one forward and one backward crossing, and its errors are then exactly the errors of backpropagation: the filled pipeline is the synchronous fixed point.

\subsection{The error as the momentum of the activation}
\label{section_canonical_structure}
At finite $c$, the forward transport \eqref{finite_c_forward} replaces the depth constraint, and the action acquires a term of first order in the time derivatives of the activations,
\begin{equation}
\label{doubled_lagrangian}
\Lambda = \Lambda_\infty + \frac1c\int_0^D\mathcal{K}_t\,d\tau .
\end{equation}
This term can pair the fields of the two copies in two ways,
\begin{equation}
\label{pairings}
\begin{split}
\mathcal{K}_t^{\rm direct} &= \bm\delta_a\cdot\partial_t\mathbf{h}^a = \bm\delta_1\cdot\partial_t\mathbf{h}_1 - \bm\delta_2\cdot\partial_t\mathbf{h}_2,\\
\mathcal{K}_t^{\rm crossed} &= \bm\delta_a\cdot\partial_t\mathbf{h}^{\dagger a} = \bm\delta_1\cdot\partial_t\mathbf{h}_2 - \bm\delta_2\cdot\partial_t\mathbf{h}_1 .
\end{split}
\end{equation}
In the direct pairing, the error of each copy is the momentum of its own activation. In the crossed pairing, it is the momentum of the activation of the other copy, the same structure $\dot x^ax^\dagger_a$ as the damping term of the parameters. On the physical limit, which is invariant, both pairings give the damped equation $\mu\,\partial_t^2W + \gamma\,\partial_tW = -(\partial\mathbf{f}/\partial W)^T\bm\delta$ and the forward transport \eqref{finite_c_forward}. They differ in the errors:
\begin{equation}
\label{pairing_errors}
\begin{split}
\text{direct: }&(\partial_t + c\,\partial_\tau)\,\bm\delta = -c\,J^T\bm\delta,\\
\text{crossed: }&(\partial_t - c\,\partial_\tau)\,\bm\delta = c\,J^T\bm\delta ,
\end{split}
\end{equation}
with $\bm\delta(D) = \partial\ell/\partial\mathbf{h}(D)$. The direct pairing gives the advanced errors of the exact gradient through training time; the crossed pairing gives exactly the master system \eqref{finite_c_system} of pipelined training, with the damping of any optimizer of Ref.~\cite{holanda2026backpropagation}. A single copy, without doubling, behaves like the direct pairing: its error is the momentum of its own activation and obeys the time-reversed, adjoint transport, so that the causal algorithm requires the doubling even without damping. In the variables $\mathbf{h}_\pm$, $\bm\delta_\pm$ (means and differences of the copies), $\mathcal{K}_t^{\rm direct} = \bm\delta_-\cdot\partial_t\mathbf{h}_+ + \bm\delta_+\cdot\partial_t\mathbf{h}_-$ and $\mathcal{K}_t^{\rm crossed} = \bm\delta_-\cdot\partial_t\mathbf{h}_+ - \bm\delta_+\cdot\partial_t\mathbf{h}_-$. The physical error $\bm\delta_+$ is conjugate to $\mathbf{h}_-$, and the crossed pairing reverses the sign of that single term. The difference of the two pairings, $-(2/c)\,\bm\delta_+\cdot\partial_t\mathbf{h}_-$, is a nonconservative potential in the sense of Galley~\cite{galley2013classical}: it is odd under the exchange of the copies, vanishes in the physical limit, and is not the difference of two single-copy terms. As $c\to\infty$ the term $\mathcal{K}_t/c$ drops out, the two pairings coincide, and the action \eqref{doubled_lagrangian} becomes the action \eqref{doubled_action_cinf} of the Coulomb limit, for which the exact gradient and the algorithm are the same.

The momenta of the parameters are unchanged. Those of the activations are now fixed by the errors, $\bm\pi_1 = \bm\delta_1/c$, $\bm\pi_2 = -\bm\delta_2/c$ for the direct pairing and $\bm\pi_1 = -\bm\delta_2/c$, $\bm\pi_2 = \bm\delta_1/c$ for the crossed pairing, and those of the errors still vanish. Since $\mathcal{K}_t$ is linear in the velocities, it drops out of the Legendre transform \eqref{doubled_legendre}: the Hamiltonian is $\mathcal{H}_\infty$ of equation \eqref{doubled_hamiltonian} for both pairings and for every $c$. What changes is the constraint structure. The constraints $\bm\pi_i \mp \bm\delta_j/c$ and $\bm\rho_i$ are second class without secondary constraints, and their Dirac bracket, equivalently the Faddeev--Jackiw symplectic form of the first-order term, makes the activations and the errors canonical pairs,
\begin{equation}
\label{bracket_pairings}
\big\{h_i(\tau, t), \delta_j(\tau', t)\big\}_D = c\,S_{ij}\,\delta(\tau - \tau'),
\end{equation}
with $S = \eta$ for the direct pairing and $S = X\eta$ for the crossed pairing, while $\{W_i, P_j\}_D = \delta_{ij}\,\delta(\tau - \tau')$ and all other brackets among $W$, $P$, $\mathbf{h}$ and $\bm\delta$ vanish. Hamilton's equations, $\partial_th_i = c\,S_{ij}\,\delta\mathcal{H}_\infty/\delta\delta_j$ and $\partial_t\delta_j = -c\,S_{ij}\,\delta\mathcal{H}_\infty/\delta h_i$, then give in the physical limit the damped weight equation, the forward transport and, for the direct and the crossed pairing respectively, the advanced and the retarded error transport \eqref{pairing_errors}: the Euler--Lagrange equations of the action \eqref{doubled_lagrangian}. The boundary terms at the output are those of the Coulomb limit. The limit $c\to\infty$ of the brackets is singular: the local bracket $c\,\delta(\tau - \tau')$ of the independent fields $\mathbf{h}$ and $\bm\delta$ diverges, while their equations of motion, multiplied by $1/c$, become the constraints of Sec.~\ref{section_depth_constraint}, which make them functionals of $W$ with the non-local brackets \eqref{dirac_reduced_cinf}. The two pairings share the action up to a nonconservative potential and share the Hamiltonian, and differ only in the bracket of the activations and the errors. \emph{Whether training follows the exact gradient or the causal algorithm is a property of the bracket.} The direct pairing at finite $c$ is not a training dynamics: its errors need data at the final training time, and it defines a boundary-value problem in $t$ rather than an initial-value problem.

This selects the pairing. In the closed-time-path formulation~\cite{galley2013classical, galley2014principle}, the variations of both copies vanish at the initial time, and at the final time the copies agree, $\mathbf{h}_- = 0$, and so do their momenta, $\bm\pi_- = 0$. For the general pairing
\begin{equation}
\label{general_pairing}
\mathcal{K}_t = \bm\delta_1\cdot(u\,\partial_t\mathbf{h}_1 + v\,\partial_t\mathbf{h}_2) - \bm\delta_2\cdot(u\,\partial_t\mathbf{h}_2 + v\,\partial_t\mathbf{h}_1),
\end{equation}
of which the direct and the crossed pairing are $(u, v) = (1, 0)$ and $(0, 1)$, these conditions make the endpoint terms of the variation vanish for every $(u, v)$: the condition on the momenta is $\bm\delta_- = 0$, and the physical error is not fixed at the final time. The endpoint conditions thus admit every pairing; what selects one is that the closed-time-path action defines an initial-value problem. In the physical limit, the pairing \eqref{general_pairing} gives activations that move forward at the speed $c/(u + v)$ and errors that move back along depth at the speed $c_b = c/(v - u)$, and, discretized in depth, the flow of the errors has the eigenvalues $-c_b/\Delta\tau$. For $v > u$, the errors relax forward in training time onto the natural boundary condition at the output. For $v < u$, they grow at the rate $|c_b|/\Delta\tau$, which diverges in the continuum: the problem is ill-posed forward in time, and the errors need data at the final time, as in backpropagation through time, the boundary-value problem of the undoubled action. \emph{Causality follows from the closed-time-path structure together with the boundary condition at the output}, and the crossed pairing is the only causal pairing in which the errors travel at the speed $c$ of the activations; the others transport the errors at a different speed. The complete brackets, including gradient flow and time-dependent coefficients, are listed in Appendix~\ref{appendix_brackets}.

\subsection{Stability: the delayed neural tangent kernel}
\label{section_stability}
At finite $c$, a change of the parameters at depth $\sigma$ reaches the output after the time $(D - \sigma)/c$, and the error it causes returns to depth $\tau$ after a further $(D - \tau)/c$. We linearize the system \eqref{finite_c_system} near an interpolating minimum $W^*$, at which $\partial\ell/\partial\mathbf{h}(D) = 0$ and hence $\bm\delta^* = 0$, for any width and any block. With $u = W - W^*$, the perturbation of the activations is transported forward along the characteristics, the error is first order and is transported back, and only the derivative $B = \partial\mathbf{f}/\partial W$ at the minimum enters the force. Integrating along the characteristics gives a closed equation for the parameters,
\begin{equation}
\label{delayed_ntk}
\begin{split}
\gamma\,\partial_tu(\tau, t) = {}&-\mathcal{J}(\tau)^TH\\
&\times\int_0^D\mathcal{J}(\sigma)\,u\Big(\sigma, t - \frac{2D - \tau - \sigma}{c}\Big)\,d\sigma ,
\end{split}
\end{equation}
where $\mathcal{J}(\sigma) = \Phi(D, \sigma)B(\sigma) = \delta\mathbf{h}(D)/\delta W(\sigma)$ is the sensitivity of the output to the parameters at depth $\sigma$, $\Phi$ is the transition matrix of the linearized forward equation, and $H$ is the Hessian of the cost at the output (with several samples, the outputs of all samples are stacked). In the Coulomb limit, equation \eqref{delayed_ntk} is the linearized gradient flow of the output with the neural tangent kernel~\cite{jacot2018neural}, $K = \int_0^D\mathcal{N}(\sigma)\,d\sigma$, $\mathcal{N} = \mathcal{J}\mathcal{J}^T$, where $\mathcal{N}(\sigma)$ is the contribution of depth $\sigma$ to the kernel. At finite $c$, each depth contributes with the delay of its round trip. Perturbations with $\int\mathcal{J}u\,d\sigma = 0$, which do not move the output, are neutral. The others are modes $u \propto e^{st - s(D - \tau)/c}\mathcal{J}(\tau)^TH\mathbf{a}$, with
\begin{equation}
\label{characteristic_ntk}
\begin{split}
&\det\big[\gamma s\,\mathbb{I} + M(s)H\big] = 0,\\
&M(s) = \int_0^D\mathcal{N}(\sigma)\,e^{-2s(D - \sigma)/c}\,d\sigma ,
\end{split}
\end{equation}
an equation in the space of the outputs, whose dimension does not grow with the depth or the width of the network. For large $c$, $M(s) = K - (2s/c)K_1 + (2s^2/c^2)K_2 + O(c^{-3})$ with the moments $K_p = \int_0^D(D - \sigma)^p\,\mathcal{N}(\sigma)\,d\sigma$, so that the output relaxes like a damped oscillator,
\begin{equation}
\label{delay_inertia_general}
\det\Big[\frac{2}{c^2}K_2H\,s^2 + \Big(\gamma\,\mathbb{I} - \frac2cK_1H\Big)s + KH\Big] = 0 .
\end{equation}
\emph{The delay acts as an inertia}, set by the second moment of the depth profile of the kernel, the spread of the kernel away from the output, and it lowers the damping by the first moment. That stale gradients act as momentum, and that delays require smaller learning rates, is known for asynchronous and pipelined training~\cite{mitliagkas2016asynchrony, yang2019pipemare, stich2020error}; for a single delay, the threshold learning rate of linearized asynchronous training is inversely proportional to the delay~\cite{giladi2019stability}, and delays can destabilize gradient flow through a Hopf bifurcation~\cite{janson2021delay}; equations \eqref{characteristic_ntk} and \eqref{delay_inertia_general} give the inertia and the threshold for a deep network in terms of its kernel.

The stability of training depends on how the kernel is distributed over depth. If $\mathcal{N}(\sigma) = \rho(\sigma)K$ with $\int\rho\,d\sigma = 1$, which holds for a single output, each eigenvalue $\kappa$ of $KH$ gives the scalar equation $\gamma s + \kappa\int\rho(\sigma)e^{-2s(D - \sigma)/c}d\sigma = 0$. With $z = sT$, the round-trip time $T = 2D/c$ and $G = \kappa T/\gamma$, a profile concentrated at the output gives $z = -G$, which is always stable; a uniform profile gives $z^2 + G(1 - e^{-z}) = 0$, which loses stability at $G = \pi^2/2$; and a profile concentrated at the input gives $z + Ge^{-z} = 0$, which loses stability at $G = \pi/2$. This is the continuum limit of delayed gradient descent: in the pipeline, a kernel at a single layer, whose round trip takes $\tau$ steps, gives $\Lambda^{\tau+1} - \Lambda^\tau + g = 0$ for the multiplier $\Lambda$ of a step, with $g$ the product of the learning rate and the eigenvalue of the kernel; its threshold $g = 2\sin[\pi/(4\tau + 2)]$ was found for asynchronous training in Ref.~\cite{giladi2019stability}, and $\tau g \to \pi/2$ for long delays. With the inertia of the heavy ball, $\gamma s$ becomes $\mu s^2 + \gamma s$, and the input-concentrated profile gives $mz^2 + z + Ge^{-z} = 0$ with $m = \mu/(\gamma T)$, whose threshold falls from $\pi/2$ to $1$ as $m$ grows: momentum shrinks the stable range under a delay, as observed for asynchronous training~\cite{giladi2019stability}. Parameters near the input, whose signals travel farthest, destabilize training; parameters near the output do not. For any profile the input-concentrated case is the worst, and the following bound holds for every network. Let $\kappa_{\max}$ be the largest eigenvalue of $KH$. \emph{If}
\begin{equation}
\label{stability_bound}
G = \frac{2D\kappa_{\max}}{\gamma c} < \frac\pi2,\qquad\text{that is,}\qquad \lambda < \frac{\pi}{4D\kappa_{\max}}
\end{equation}
with $c = 1/(\gamma\lambda)$ from equation \eqref{c_learning_rate}, \emph{equation \eqref{characteristic_ntk} has no root with} $\operatorname{Re}s \ge 0$ other than the neutral modes. Proof: for a root $s \neq 0$ with eigenvector $\mathbf{w}$, let $\mathbf{y} = H^{1/2}\mathbf{w}$ (nonzero, since otherwise $s\mathbf{w} = 0$). Then $\gamma s|\mathbf{y}|^2 + \int_0^Dq(\sigma)e^{-2s(D - \sigma)/c}d\sigma = 0$ with $q = \mathbf{y}^\dagger H^{1/2}\mathcal{N}H^{1/2}\mathbf{y} \ge 0$ and $\int q\,d\sigma \le \kappa_{\max}|\mathbf{y}|^2$. If $\operatorname{Re}s \ge 0$, the imaginary part gives $\gamma|\operatorname{Im}s| \le \kappa_{\max}$, so every phase $2|\operatorname{Im}s|(D - \sigma)/c$ is at most $G < \pi/2$, and the real part of the left-hand side is positive, a contradiction. The bound uses only that the kernel is positive semidefinite and that the delays are at most $T$: the round trip of the signals must be shorter than $\pi/2$ times the fastest relaxation time $\gamma/\kappa_{\max}$ of the output. It is sharp for a kernel concentrated at the input, and for other profiles the threshold is higher. We test these results in Sec.~\ref{section_stability_test}.

\subsection{Look-ahead and weight prediction}
\label{section_lookahead}
Nesterov's look-ahead extrapolates the parameters by one step, $Y = W + h\,\partial_tW$ (Sec.~\ref{section_nesterov}). In a pipeline the step is $h = \Delta\tau/c$, so this look-ahead vanishes in the continuum limit and cannot compensate a delay of the order of the round trip $T = 2D/c$. Pipelined training instead predicts the parameters over a longer time, from the last update or the momentum~\cite{chen2018efficient, kosson2020pipelined}, and asynchronous training compensates the delay by a Taylor correction~\cite{zheng2017asynchronous}. We let the passes at depth $\sigma$ use the predicted parameters $Y = W + \varepsilon(\sigma)\,\partial_tW$, with a look-ahead time $\varepsilon$; near a minimum only the forward pass matters, since the error vanishes there. These are the look-ahead fields of Sec.~\ref{section_nesterov} with the look-ahead time $\varepsilon(\sigma)$, which give the look-ahead equations from the action, for the heavy ball and for gradient descent, in the Coulomb limit and at finite $c$, with the transport driven by $Y$. Linearized at an interpolating minimum, the transport is driven by $Y$, and for a mode $Y = (1 + \varepsilon s)u$, so that the delayed kernel \eqref{characteristic_ntk} becomes
\begin{equation}
\label{lookahead_kernel}
M_\varepsilon(s) = \int_0^D\mathcal{N}(\sigma)\,\big(1 + \varepsilon(\sigma)s\big)\,e^{-2s(D - \sigma)/c}\,d\sigma .
\end{equation}
We consider a constant look-ahead $\varepsilon = aT$ and a proportional one, $\varepsilon(\sigma) = a\cdot2(D - \sigma)/c$, in which each layer predicts a fraction $a$ of its own round trip; $a = 1$ predicts over the full delay of each layer. For the proportional look-ahead, $M_\varepsilon = K - (1 - a)(2s/c)K_1 + (1 - 2a)(2s^2/c^2)K_2 + O(c^{-3})$: the prediction over the full delay cancels the loss of damping in equation \eqref{delay_inertia_general}, but it reverses the sign of the inertia, and $a = 1/2$ removes the inertia instead. For a kernel at a single delay, $z + G(1 + bz)e^{-z} = 0$ with $b = \varepsilon/T$, and $z = iy$ is a root if $\cos y + by\sin y = 0$ and $G = y/(\sin y - by\cos y)$. The threshold rises from $\pi/2$ to $1.819$ at $b = 1/4$ and falls to $0.942$ at $b = 1$: \emph{a partial look-ahead stabilizes gradient descent, and a prediction over the full delay overcompensates the delay}. For the proportional look-ahead, a kernel at the delay $uT$ has the threshold $G_c(a)/u$, so that the kernel at the input remains the worst case among single delays. For a uniform profile, a constant look-ahead raises the threshold from $\pi^2/2$ to $12.1$ at $a = 1/4$.

The pipeline estimates the velocity from the last update, $Y_k = W_k + e_k(W_k - W_k^{\rm old})$ with $e_k = \varepsilon_k/h$, and this estimate adds a mode that the continuum does not contain: the alternating mode $\Lambda = -1$ of the step multiplier, which loses stability when
\begin{equation}
\label{alternating_mode}
\det\Big[-2\,\mathbb{I} + \frac{h}{\Delta\tau}\sum_k(1 + 2e_k)\,J_kJ_k^TH\Big] = 0 ,
\end{equation}
with $J_k = \partial\mathbf{h}_N/\partial W_k$. For a constant look-ahead this gives $G = 1/[a(1 + 1/(4aN))] \to 1/a$, for any kernel. The prediction amplifies the alternating mode by the factor $1 + 2e_k$, which grows with the number of steps over which it extrapolates, and limits large look-aheads. The stable range is therefore bounded by the continuum threshold of the kernel \eqref{lookahead_kernel}, which a moderate look-ahead raises, and by the alternating mode, which a large one destabilizes. For a kernel at a single layer with a round trip of $\tau$ steps, the discrete equation is $\Lambda^{\tau+2} - \Lambda^{\tau+1} + g(1 + e)\Lambda - ge = 0$, the characteristic polynomial of linear weight prediction without momentum, with the horizon $e$~\cite{kosson2020pipelined}; for long delays its thresholds tend to those of the kernel \eqref{lookahead_kernel}. The gain of a partial look-ahead requires a round trip of several steps: for a delay of one step every look-ahead lowers the threshold, a prediction over the full delay from $1$ to $(\sqrt5 - 1)/2$, consistent with the small loss of stability that Ref.~\cite{kosson2020pipelined} reports for weight prediction at small momentum, while the best gain, about $2\%$, $6\%$ and $10\%$ for delays of $2$, $4$ and $8$ steps, approaches the continuum value of $16\%$ for long delays.

\emph{Momentum.} With the inertia of the heavy ball, a look-ahead over the time $\mu/\gamma$ of the momentum removes the inertia, and not only near a minimum. If both passes use $Y = W + (\mu/\gamma)\,\partial_tW$, the weight equation $\mu\,\partial_t^2W + \gamma\,\partial_tW = -F[Y]$ becomes $\gamma\,\partial_tY = -F[Y]$, and the parameters follow through $(\mu/\gamma)\,\partial_tW + W = Y$: the predicted parameters obey the delayed gradient flow and the parameters are their low-pass filter, for any network, any force $F$ and any delays. In the action, this is the bracket of gradient flow that the look-ahead fields acquire at $\varepsilon = \mu/\gamma$ (Sec.~\ref{section_nesterov}). The pipeline has the same identity exactly. With both passes at $u = W + e(W - W^{\rm old})$ and $e = \beta/(1 - \beta)$ at every layer, the heavy ball $W \to W + \beta(W - W^{\rm old}) - \alpha g(u)$ gives $u \to u - \alpha g(u)/(1 - \beta)$ for any force $g$, and $W = (1 - \beta)u + \beta W^{\rm old}$: the pipelined heavy ball is pipelined gradient descent at the effective rate $\alpha/(1 - \beta)$ with an exponential moving average of the parameters. Without delays, this is the equivalence of momentum and weight averaging~\cite{defazio2024road, morwani2025connections}; for a single delay and a quadratic loss, it follows from the equivalence of weight prediction and spike compensation in Ref.~\cite{kosson2020pipelined}, whose compensation then has no velocity term. Near a minimum, the characteristic equation becomes $(1 + \mu s/\gamma)\det[\gamma s\,\mathbb{I} + M(s)H] = 0$, the delayed gradient flow with an additional decaying mode, and in the pipeline it factors into $\Lambda - \beta$ and the equation of pipelined gradient descent at the effective rate. Nesterov's look-ahead, $e = \beta$, is shorter by the factor $1 - \beta$ and does not reduce the heavy ball; in the deep network it raises the threshold by about $1\%$ (Sec.~\ref{section_stability_test}), as Ref.~\cite{kosson2020pipelined} finds that Nesterov's method is only marginally better than momentum for delays longer than one step. Beyond the momentum time, the best look-ahead grows with the momentum. For a kernel at a single delay, $mz^2 + z + G(1 + bz)e^{-z} = 0$ with $m = \mu/(\gamma T)$: $b = m$ gives $G_c = \pi/2$ for every $m$, the optimal $b$ grows from $0.25$ at $m = 0$ to $0.95$ at $m = 2$, and a prediction over the full delay beats no prediction for $m > 0.504$ ($G_c = 2.166$ against $1.075$ at $m = 2$). In the moment expansion, the physical inertia offsets the negative inertia of the prediction: the proportional look-ahead with $a = 1$ leaves the damping $\gamma$ and the inertia $\mu - (2/c^2)K_2H$. This agrees with Ref.~\cite{kosson2020pipelined}, which finds that with momentum weight prediction enlarges the stable region and that horizons beyond the delay converge fastest on a quadratic. In its setting, a single delay of $D$ steps and a condition number of $10^3$, with the step and the momentum tuned, the convergence time falls by a factor of about $2$ from $T = D$ to $T = 1.25D$ ($607$ to $284$ steps for $D = 5$, $1168$ to $663$ for $D = 10$) and by less than $2\%$ more up to $T = 2.5D$.

\section{Numerical tests}
\label{section_numerical_tests}

\begin{figure*}
\includegraphics[width=\textwidth]{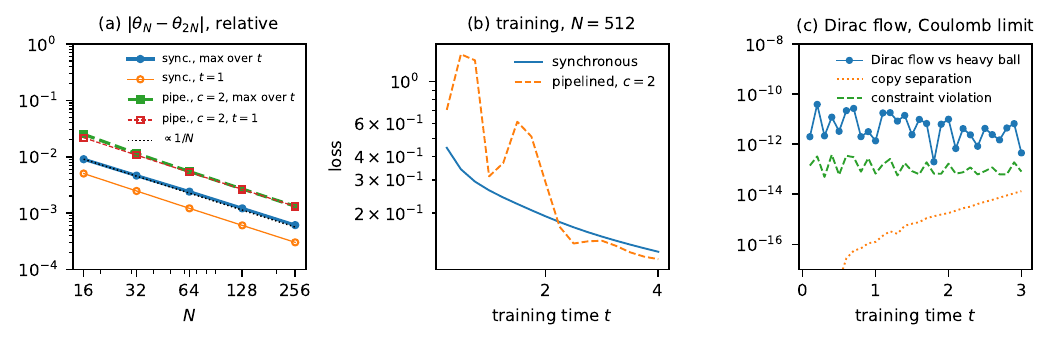}
\caption{\label{fig_master_deepnet}Deep residual networks with blocks $U\tanh(V\mathbf{h} + \mathbf{b})$. (a) Relative difference between the parameter fields of networks with $N$ and $2N$ layers, maximum over the training interval (filled) and at $t = 1$ (open), for synchronous gradient flow and for pipelined training at $c = 2$. (b) Loss for $N = 512$. (c) Dirac flow of the doubled action in the Coulomb limit, integrated from the physical limit for a network with $N = 4$: deviation of the copy average from ordinary heavy-ball training with gradients by reverse-mode differentiation, separation of the copies, and violation of the constraints.}
\end{figure*}

\begin{figure*}
\includegraphics[width=\textwidth]{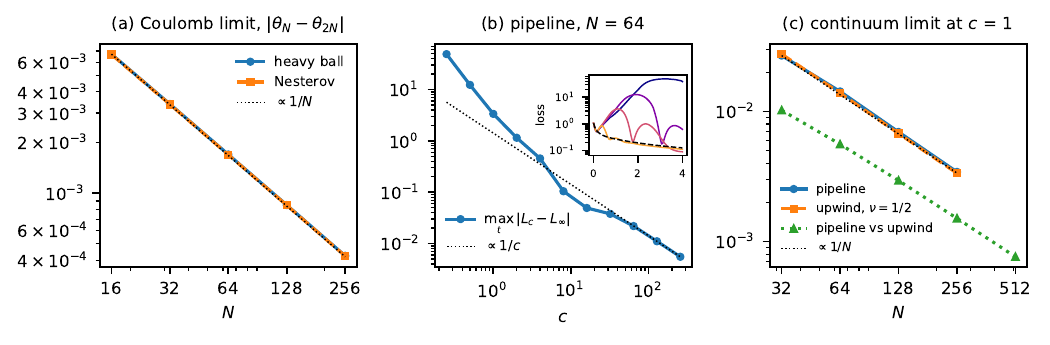}
\caption{\label{fig_master_deeptests}Deep residual networks of Fig.~\ref{fig_master_deepnet}. (a) Heavy ball and Nesterov's method ($\mu = 1$, $\gamma = 2$) in the Coulomb limit: relative difference between the parameter fields of networks with $N$ and $2N$ layers, maximum over $t \le 4$. (b) Pipelined gradient descent with $N = 64$: largest deviation of the loss from synchronous gradient flow against $c$, and (inset) the loss for $c = 1/4$, $1/2$, $1$ and $4$ and for synchronous training (dashed). (c) Continuum limit at $c = 1$ of the pipeline and of an upwind discretization with Courant number $1/2$, and the difference between the two.}
\end{figure*}

\begin{figure*}
\includegraphics[width=\textwidth]{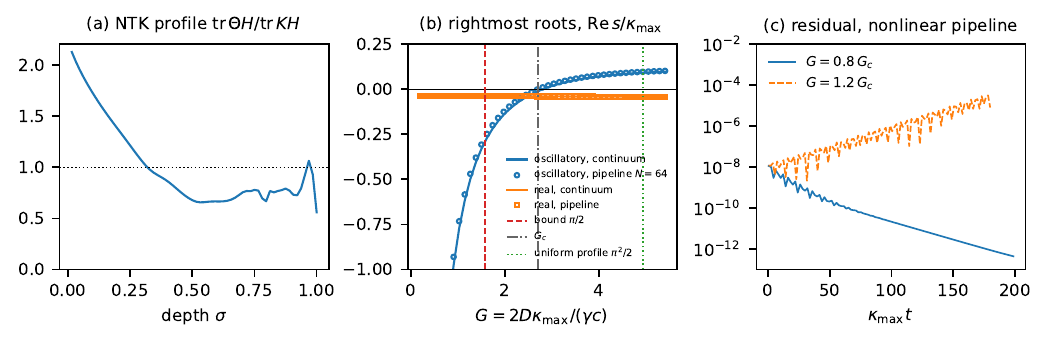}
\caption{\label{fig_master_delay}Stability of pipelined training of a deep residual network with blocks $U\tanh(V\mathbf{h} + \mathbf{b})$ near an interpolating minimum ($N = 64$). (a) Depth profile of the neural tangent kernel, $\operatorname{tr}\mathcal{N}(\sigma)H/\operatorname{tr}KH$, and the uniform profile (dotted). (b) Real parts of the rightmost real root and of the rightmost complex root of equation \eqref{characteristic_ntk} (lines) and of the exact characteristic equation of the pipeline (markers) against $G = 2D\kappa_{\max}/(\gamma c)$, with the bound \eqref{stability_bound}, the threshold $G_c$ and the threshold of a uniform profile. (c) Norm of the residual of the nonlinear pipeline started near the minimum, below and above the threshold.}
\end{figure*}

\begin{figure*}
\includegraphics[width=\textwidth]{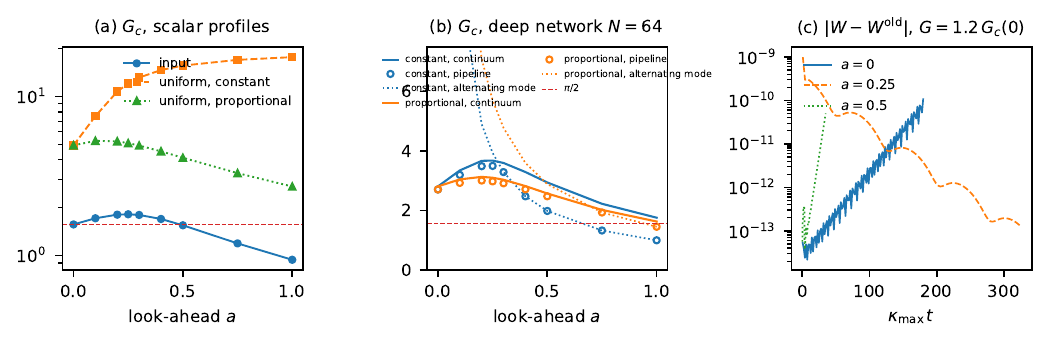}
\caption{\label{fig_master_lookahead}Look-ahead in pipelined training near a minimum. (a) Thresholds $G_c$ of the kernel \eqref{lookahead_kernel} against the look-ahead $a$, for a kernel concentrated at the input and for a uniform profile with a constant and a proportional look-ahead; the dashed line is $\pi/2$. (b) Thresholds of the deep network of Fig.~\ref{fig_master_delay} ($N = 64$) for the constant and the proportional look-ahead: continuum kernel (lines), pipeline (markers) and the alternating mode \eqref{alternating_mode} (dotted). (c) Norm of the velocity estimate $W - W^{\rm old}$ of the nonlinear pipeline started near the minimum at $G = 1.2\,G_c$, without look-ahead and with the constant look-ahead $a = 1/4$ and $a = 1/2$.}
\end{figure*}

\subsection{Network and data}
We train residual networks with blocks $\mathbf{f} = U\tanh(V\mathbf{h} + \mathbf{b})$ of width $4$ and hidden width $8$ on $16$ samples, with inputs $(x_1, x_2)$ drawn uniformly from $[-1, 1]^2$ in the first two components of $\mathbf{h}_0$, targets $\sin(\pi x_1)\,x_2 + 0.3$ for the first component of the output, the squared-error cost and $D = 1$. The initial parameter fields $U(\tau)$, $V(\tau)$ and $\mathbf{b}(\tau)$ are random combinations of $1$, $\sin\pi\tau$ and $\cos2\pi\tau$, so that networks of different depth sample the same continuum fields. We compare the parameter fields of networks with $N$ and $2N$ layers at their common depths, in the norm $\|W\| = (\int_0^D|W|^2d\tau)^{1/2}$ of all parameter fields, relative to the norm of the fields.

\subsection{Continuum limit}
\emph{Coulomb limit.} Gradient descent with the learning rate \eqref{depth_scaled_rates}, $\gamma = 1$ and the step $h = 0.01$, up to $T = 4$, converges to the master system \eqref{master_system} at first order in $\Delta\tau$ (Fig.~\ref{fig_master_deepnet}a): the largest difference over the training interval is $9.0\times10^{-3}$ at $N = 16$ and $6.1\times10^{-4}$ at $N = 256$, with ratios of successive differences $1.95$, $1.93$, $1.98$ and $1.98$ ($2.04$ to $2.00$ at $t = 1$). The heavy ball in the Polyak form with the rates \eqref{depth_scaled_rates}, $\mu = 1$ and $\gamma = 2$, started from rest, converges in the same way (Fig.~\ref{fig_master_deeptests}a), from $6.7\times10^{-3}$ at $N = 16$ to $4.2\times10^{-4}$ at $N = 256$, with ratios $1.98$, $1.99$, $2.00$ and $2.00$, and so does Nesterov's method with the same rates, whose differences agree with those of the heavy ball to within $1\%$ at this step.

\emph{Finite $c$.} The pipeline \eqref{pipeline_update} at $c = 2$, started from the filled pipeline, converges to the system \eqref{finite_c_system} at first order (Fig.~\ref{fig_master_deepnet}a), from $2.5\times10^{-2}$ at $N = 16$ to $1.3\times10^{-3}$ at $N = 256$, with ratios $2.19$, $2.09$, $2.05$ and $2.02$. Early in training its loss oscillates (Fig.~\ref{fig_master_deepnet}b). The pipelined heavy ball, with $\mu = 1$, $\gamma = 2$ and the rates \eqref{depth_scaled_rates} at $h = \Delta\tau/c$, started from rest, converges at first order as well, from $2.0\times10^{-2}$ at $N = 16$ to $2.0\times10^{-3}$ at $N = 128$, with ratios $2.28$, $2.15$ and $2.07$. An independent discretization of the system \eqref{finite_c_system}, an upwind scheme with Courant number $1/2$, $\mathbf{h}_k \leftarrow \frac12\mathbf{h}_k + \frac12[\mathbf{h}_{k-1} + \Delta\tau\,\mathbf{f}(W_k, \mathbf{h}_{k-1})]$, likewise for the errors, and $h = \Delta\tau/(2c)$, converges in the same way at $c = 1$ (ratios $1.99$, $2.04$ and $2.02$ for $N = 32$ to $256$, against $1.89$, $2.05$ and $2.03$ for the pipeline), and the two discretizations approach each other at first order, from $1.0\times10^{-2}$ at $N = 32$ to $7.6\times10^{-4}$ at $N = 512$ (Fig.~\ref{fig_master_deeptests}c). At $N = 64$, the deviation of the loss from synchronous gradient flow, integrated accurately at the same $N$, falls like $1/c$ once the round trip $2D/c$ is short compared with the relaxation time of the loss: its maximum over $t \le 4$ is $2.2\times10^{-2}$, $1.1\times10^{-2}$ and $5.5\times10^{-3}$ for $c = 64$, $128$ and $256$ (Fig.~\ref{fig_master_deeptests}b). For $c \le 8$ the loss is not monotone, at $c = 1/4$ training diverges, and for $1 \le c \le 8$ the loss at $T = 4$ is below that of synchronous training ($0.088$ at $c = 1$ against $0.124$).

\subsection{Backpropagation from the Dirac flow}
\label{section_dirac_test}
We test the Dirac evolution of Secs.~\ref{section_depth_constraint} and~\ref{section_finite_c} as in Ref.~\cite{holanda2026backpropagation}, with a generic Dirac flow, $\dot z = \Omega\nabla H - \Omega J^T\mathcal{M}^{-1}J\,\Omega\nabla H$ with $J = \partial\Theta/\partial z$ and $\mathcal{M} = J\Omega J^T$ for the constraints $\Theta$, in which the Hamiltonian, the constraints and all their derivatives are obtained by automatic differentiation and no backward pass is coded. For residual blocks $U\tanh(V\mathbf{h} + \mathbf{b})$ ($N = 4$, width and hidden width $3$, four samples, $\mu = 1$, $\gamma = 2$), at a point of the constraint surface of the Coulomb limit with the two copies $0.35$ apart, the Dirac evolution of the momenta equals the exact gradient computed by reverse-mode differentiation to $3\times10^{-17}$, on a scale of $0.5$, and that of the activations equals the forward-mode Jacobian--vector product to $2\times10^{-16}$; the Poisson flow alone leaves the activations frozen. At finite $c$ ($c = 1.5$), in the physical limit, the Dirac flow of the crossed pairing equals the pipeline to $2\times10^{-15}$. Integrated from the physical limit, the Dirac flow of the Coulomb limit follows ordinary heavy-ball training to $4\times10^{-11}$ up to $t = 3$, with the constraints satisfied to $4\times10^{-13}$ and the copies separated by $10^{-14}$ (Fig.~\ref{fig_master_deepnet}c).

As in Ref.~\cite{holanda2026backpropagation}, the physical limit is invariant but unstable, since the mirror copy is anti-damped; the integration starts on it, and the separation of the copies stays at round-off level over the interval shown.

\subsection{Stability near a minimum}
\label{section_stability_test}
We test the stability analysis of Sec.~\ref{section_stability} on the same network with $4$ samples, at an interpolating minimum found by Gauss--Newton iteration (residual below $10^{-15}$). For the discrete pipeline \eqref{pipeline_update}, the same reduction gives the exact characteristic equation $\det[(\Lambda - 1)\mathbb{I} + (h/\Delta\tau)\sum_k\Lambda^{-2(N - k + 1)}J_kJ_k^TH] = 0$ for the multiplier $\Lambda = e^{sh}$ of one step, with $J_k = \partial\mathbf{h}_N/\partial W_k$. For $N = 8$, its $68$ roots are eigenvalues of the Jacobian of the full update map, of dimension $832$, to $7\times10^{-15}$, and every other eigenvalue is neutral ($\Lambda = 1$) or belongs to the pure transport of signals ($|\Lambda| \le 0.05$). The kernel of the trained network is weighted toward the input (Fig.~\ref{fig_master_delay}a), and the minimum loses stability at $G = 2.673$, $2.706$ and $2.724$ in the pipeline for $N = 32$, $64$ and $128$, and at $G = 2.881$, $2.809$ and $2.774$ from equation \eqref{characteristic_ntk} on the same minima: the two approach each other at first order in $\Delta\tau$, between the bound $\pi/2$ and the uniform value $\pi^2/2$ (Fig.~\ref{fig_master_delay}b). Below the threshold the slowest mode is real, above it a complex pair grows, with $\operatorname{Im}s\,T = 2.22$ at onset. Started from a perturbation of size $10^{-6}$, the nonlinear pipeline with $N = 64$ follows the rightmost root: its residual decays at the rate $0.281$ at $G = 0.8\,G_c$ (predicted $0.283$) and grows at the rate $0.3125$ at $G = 1.2\,G_c$ (predicted $0.3130$), with oscillations (Fig.~\ref{fig_master_delay}c).

\emph{Look-ahead.} With the look-ahead of Sec.~\ref{section_lookahead}, the exact characteristic equation of the pipeline is $\det[(\Lambda - 1)\mathbb{I} + (h/\Delta\tau)\sum_k(1 + e_k(1 - \Lambda^{-1}))\Lambda^{-2(N - k + 1)}J_kJ_k^TH] = 0$; for $N = 8$ its $72$ roots are eigenvalues of the Jacobian of the full update map with prediction, of dimension $1408$, to $3\times10^{-14}$, for both families. At $N = 64$ (Fig.~\ref{fig_master_lookahead}b), a constant look-ahead $a = 1/4$ raises the threshold of the pipeline from $2.706$ to $3.499$, by $29\%$, and the proportional look-ahead $a = 1/5$ to $3.006$. The low-frequency roots of the pipeline follow the kernel \eqref{lookahead_kernel}, whose thresholds lie $3\%$ to $5\%$ higher at this $N$, as without look-ahead. For larger look-aheads the pipeline loses stability through the alternating mode, at the values given by equation \eqref{alternating_mode} to the digits shown: $1.9845$ for the constant $a = 1/2$, and $1.4517$ for the proportional $a = 1$, the prediction over the full delay, against $2.706$ without prediction. Started near the minimum at $G = 1.2\,G_c = 3.248$, the nonlinear pipeline without prediction diverges at the rate $0.3132$ (predicted $0.3130$), with the constant look-ahead $a = 1/4$ it converges at the rate $0.1722$ (predicted $0.1722$), and with $a = 1/2$ it diverges through the alternating mode at the rate $1.6243$ (predicted $1.6256$), alternating in sign from step to step (Fig.~\ref{fig_master_lookahead}c).

\begin{figure}
\includegraphics[width=\columnwidth]{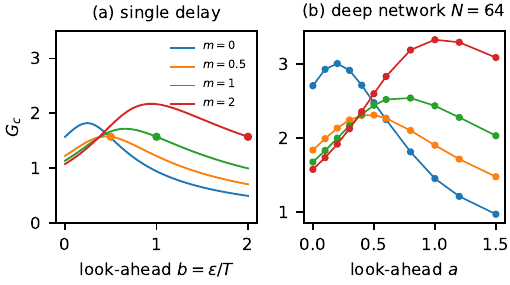}
\caption{\label{fig_master_lookahead_momentum}Look-ahead with momentum, $m = \mu/(\gamma T)$ the momentum time in round trips. (a) Thresholds of a kernel at a single delay, $mz^2 + z + G(1 + bz)e^{-z} = 0$, against $b = \varepsilon/T$; the dots mark $b = m$, where $G_c = \pi/2$. (b) Thresholds of the pipelined heavy ball on the deep network ($N = 64$) against the proportional look-ahead $a$, in the effective $G$.}
\end{figure}

\emph{Look-ahead with momentum.} For the pipelined heavy ball with prediction, the characteristic equation has $(\Lambda - 1)(\Lambda - \beta)/\Lambda$ in place of $\Lambda - 1$; for $N = 8$ its roots are eigenvalues of the full update map to $1.3\times10^{-14}$, the remaining eigenvalues being those of the neutral modes, $\Lambda = 1$ and $\Lambda = \beta$. At $N = 64$ with $\beta = 0.95$ and $e = \beta/(1 - \beta)$, its $520$ roots are $\beta$ and the roots of pipelined gradient descent at the effective rate to $3\times10^{-12}$. The reduction holds for the nonlinear pipeline as well. For $N = 16$, started from the initial fields far from the minimum, with $\beta = 0.9$, $c = 4$ and both passes at the predicted parameters, the predicted parameters of the heavy ball with $e = \beta/(1 - \beta)$ and the parameters of pipelined gradient descent at the effective rate agree to $3\times10^{-13}$ (relative) over $3000$ ticks, in which the loss falls from $0.573$ to $4\times10^{-6}$ and the parameters change by $150\%$; the activations and the errors agree to $1.3\times10^{-12}$, and the parameters equal the moving average $(1 - \beta)u + \beta W^{\rm old}$ to $2\times10^{-16}$. With Nesterov's horizon $e = \beta$, the two trajectories differ at order one. We set $\beta = 1 - 1/(2Nm)$, so that the momentum time is $m$ round trips, and give the thresholds in the effective $G = 2D\kappa_{\max}/[c(1 - \beta)]$ (Fig.~\ref{fig_master_lookahead_momentum}b). Without prediction, momentum lowers the threshold from $2.706$ to $1.836$, $1.674$ and $1.575$ for $m = 1/2$, $1$ and $2$; the look-ahead over the momentum time restores $2.706$ to all digits, while Nesterov's look-ahead gives $1.856$, $1.693$ and $1.593$; and the best proportional look-ahead grows with the momentum, from $a = 0.2$ ($3.006$) without momentum to $a \approx 0.5$ ($2.309$), $0.8$ ($2.540$) and $1$ ($3.328$). At $m = 2$, linear weight prediction over each layer's delay is the best of the family and exceeds the best threshold without momentum.

\section{Conclusion}
\label{section_conclusion}

We have shown that, in the continuum limit, the training of a deep residual network is a $1+1$ field theory, with the depth as space and the training time as time. The master system \eqref{master_system}, the continuum limit of gradient descent and of the heavy ball, is local in both variables for any width and any form of the blocks, and the non-locality of backpropagation is carried by the forward and adjoint equations, which hold at each training time. Its continuum limit requires a learning rate that grows in proportion to the depth. In the Bateman--Galley doubled action, with the errors as fields, the forward pass and the backward pass of each copy are the secondary constraints of Dirac's algorithm, and the Dirac bracket reduces the theory to the phase space of the parameters, with a kernel that is retarded along depth and equal to the Jacobian of the network at the output; the physical limit of the Dirac evolution is the damped master system, for any optimizer of the heavy-ball type. A finite propagation speed $c$, which for pipelined training is the inverse of the per-parameter learning rate, turns these constraints into transport equations, the activations moving forward and the errors backward along depth, both forward in training time. Near an interpolating minimum, the delays make the neural tangent kernel a kernel in time as well as in depth: the delay acts as an inertia set by the depth profile of the kernel, and the training of any such network is stable when the round trip of the signals is shorter than $\pi/2$ times the fastest relaxation time of the output, with a threshold that is higher when the kernel is accumulated near the output. At finite $c$, the error becomes the momentum of an activation, and the choice of which activation decides the error transport: the copy's own activation gives the advanced errors of the exact gradient through training time, and the other copy's activation, the pairing of the damping term, gives the retarded errors of the algorithm. The closed-time-path structure of the action, an initial-value problem in training time with the boundary condition at the output, admits only causal pairings, and the crossed pairing is the one in which the errors travel at the speed of the activations. The two pairings share the Hamiltonian and differ only in the bracket, of which we have given the complete Poisson and Dirac form for finite and infinite $c$, with and without inertia, and with time-dependent coefficients; integrated numerically, the Dirac flow reproduces backpropagation in time and depth without a coded backward pass. Many results of optimization and machine learning, from backpropagation, the depth scaling of the learning rate and the implicit momentum of asynchronous training to the threshold learning rate of delayed gradient descent, the characteristic polynomial of weight prediction and the equivalence of momentum and weight averaging, follow from the master system as solutions and limits and agree exactly with the published formulas. The same equations give the bound on the round trip of the signals for any network, the dependence of the delays on the depth profile of the kernel and the exact reduction of pipelined momentum training, with a look-ahead over the time of the momentum, to delayed gradient descent with weight averaging. Weight prediction enters the action through look-ahead fields, for the heavy ball and, through the second-class constraints of gradient flow, for gradient descent; at the time of the momentum, the look-ahead fields of the heavy ball carry the Dirac bracket of gradient flow, so that the heavy ball contains a gradient-flow theory. The master system thus plays for residual networks the role of a field equation, with these results among its solutions and limits.
Several questions remain open. The theory covers residual networks trained by optimizers of the heavy-ball type; skip connections across several layers, adaptive optimizers, normalization layers and the noise of stochastic gradients are not yet part of it. Near a minimum there are no propagating waves along depth; a term coupling the parameters at neighboring depths, such as a depth-smoothness regularizer or partial weight sharing, would add a tension $-\kappa\,\partial_\tau^2W$ to equation \eqref{master_weights}, with shared weights, that is, a recurrent network, as $\kappa\to\infty$. The symmetries of the $1+1$ theory are of a new kind. The balances of the rescaling symmetry of homogeneous activations~\cite{du2018algorithmic, kunin2020neural} are conserved by gradient flow at each depth for any error, and hence also by the causal algorithm, while symmetries that act along depth, with a connection, are sensitive to the causal structure; conservation laws become balance laws with fluxes through the input and output layers, a finite $c$ allows transformations that mix depth and training time, and anisotropic scalings $t \sim \tau^z$ may connect with the depth-scaling rules of learning rates~\cite{yang2023tensor, bordelon2023depthwise}. Finally, physical learning machines, in which the units relax on their own time scale, as in equilibrium propagation~\cite{scellier2016equilibrium} and predictive coding~\cite{rao1999predictive, whittington2017approximation}, suggest a $1+2$ theory with a separate relaxation clock, of which the present theory and that of Ref.~\cite{scurria2026physical} would be limits.

\section*{Data availability}
The code that produces Figs.~\ref{fig_master_deepnet}--\ref{fig_master_lookahead_momentum} and the numbers in Sec.~\ref{section_numerical_tests} is openly available~\cite{holanda2026twotime}. The data are synthetic and are generated by this code.

\appendix
\section{Gradient flow}
\label{appendix_gradient_flow}

For $\mu = 0$, the Lagrangian of the parameters in equation \eqref{doubled_action_cinf}, $\frac\gamma2(\dot W_1W_2 - \dot W_2W_1)$, is linear in the velocities, and the momenta are fixed by the coordinates. Per unit depth, and in plain canonical momenta,
\begin{equation}
\label{gradient_flow_constraints}
\begin{split}
&\chi_1 = P_1 - \frac\gamma2W_2 \approx 0,\qquad \chi_2 = P_2 + \frac\gamma2W_1 \approx 0,\\
&\big\{\chi_1(\tau, t), \chi_2(\tau', t)\big\} = -\gamma\,\delta(\tau - \tau'),
\end{split}
\end{equation}
one pair of second-class constraints at each depth, as in Appendix~A of Ref.~\cite{holanda2026backpropagation}. At finite $c$, the constraints of the activations and errors, $\bm\pi_i$ minus a multiple of an error and $\bm\rho_i$, contain no parameters, so that they commute with the constraints \eqref{gradient_flow_constraints}, and the constraint matrix is block diagonal. The Dirac brackets of the two sectors then simply combine: those of the parameters are $\{W_1(\tau, t), W_2(\tau', t)\}_D = \gamma^{-1}\delta(\tau - \tau')$, $\{W_i, P_j\}_D = \frac12\delta_{ij}\,\delta(\tau - \tau')$ and $\{P_1, P_2\}_D = \frac\gamma4\,\delta(\tau - \tau')$, those of the activations and errors are given by equation \eqref{bracket_pairings}, and the two sectors are coupled only through $\mathbf{f}(W, \mathbf{h})$ in the Hamiltonian. In the physical limit, Hamilton's equations give $\gamma\,\partial_tW = -(\partial\mathbf{f}/\partial W)^T\bm\delta$ with the forward transport and, for the crossed pairing, the retarded error transport: pipelined gradient descent.

In the Coulomb limit, the depth constraint and the adjoint equation depend on the parameters and have nonvanishing brackets with the constraints \eqref{gradient_flow_constraints}, and the constraint matrix is not block diagonal. The Dirac bracket is then built in stages~\cite{henneaux1992quantization}, as in Ref.~\cite{holanda2026backpropagation}: imposing the constraints of the activations and errors first replaces the cost by the loss $L[W] = \ell(\mathbf{h}(D; W))$, and imposing the constraints \eqref{gradient_flow_constraints} afterwards gives gradient flow on the composed network, $\gamma\,\partial_tW(\tau) = -\delta L/\delta W(\tau)$, with the brackets of Appendix~\ref{appendix_brackets}. With a time-dependent $\gamma(t)$, the constraints \eqref{gradient_flow_constraints} depend explicitly on time, and the evolution contains the additional term $-\{f, \Theta\}\,\mathcal{M}^{-1}\partial\Theta/\partial t$ of Ref.~\cite{holanda2026backpropagation}.

\section{Complete brackets}
\label{appendix_brackets}

As in Ref.~\cite{holanda2026backpropagation}, we give the complete brackets. They are equal-time brackets, between fields on a slice of constant training time, with the depth as the coordinate along the slice. We write them with plain canonical momenta, which is the bracket $\Gamma = \Gamma_1 - \Gamma_2$ of Ref.~\cite{holanda2026backpropagation} in its covariant momenta. The Poisson brackets are $\{W_i(\tau, t), P_j(\tau', t)\} = \{h_i(\tau, t), \pi_j(\tau', t)\} = \{\delta_i(\tau, t), \rho_j(\tau', t)\} = \delta_{ij}\,\delta(\tau - \tau')$, with all others zero. The Dirac brackets depend on $c$ and $\mu$:
\begin{itemize}
\item Finite $c$, $\mu \neq 0$. All brackets are local: $\{W_i, P_j\}_D = \delta_{ij}\,\delta(\tau - \tau')$, the bracket \eqref{bracket_pairings} of the activations and the errors, and all other brackets among $W$, $P$, $\mathbf{h}$ and $\bm\delta$ zero.
\item Finite $c$, $\mu = 0$. The constraints of the parameters (Appendix~\ref{appendix_gradient_flow}) add $\{W_i, W_j\}_D = \gamma^{-1}(\eta X)_{ij}\,\delta(\tau - \tau')$, $\{W_i, P_j\}_D = \frac12\delta_{ij}\,\delta(\tau - \tau')$ and $\{P_1, P_2\}_D = \frac\gamma4\,\delta(\tau - \tau')$; the parameters still commute with the activations and the errors.
\item Coulomb limit, $\mu \neq 0$. The reduced phase space of Sec.~\ref{section_depth_constraint}, equations \eqref{dirac_reduced_cinf} and \eqref{retarded_kernel}, for each copy.
\item Coulomb limit, $\mu = 0$. The constraints of the activations and errors depend on the parameters and do not commute with those of the parameters, and the bracket is built in stages, as in Ref.~\cite{holanda2026backpropagation}: $\{F, G\}_D = \gamma^{-1}\int_0^D\big(\frac{\delta F}{\delta W_1}\frac{\delta G}{\delta W_2} - \frac{\delta F}{\delta W_2}\frac{\delta G}{\delta W_1}\big)d\tau''$ for functionals of $W$, $\mathbf{h}$, $\bm\delta$. The activations of the two copies then do not commute, as for Nesterov's look-ahead in Ref.~\cite{holanda2026backpropagation}.
\end{itemize}
\emph{Time-dependent coefficients.} With $\gamma(t)$ and $\mu(t)$, the Lagrangian of the parameters contains the additional terms of Ref.~\cite{holanda2026backpropagation}. For gradient flow the constraints then depend on $t$, so that $\{W_1, W_2\}_D = \delta(\tau - \tau')/\gamma(t)$, and the evolution acquires the term $-\{f, \Theta\}\,\mathcal{M}^{-1}\partial\Theta/\partial t$ of Ref.~\cite{holanda2026backpropagation}, without which the physical limit is not $\gamma(t)\,\partial_tW = -f_W^T\bm\delta$. A time-dependent speed, such as $c = m/(\gamma\lambda(t))$ for a pipeline with a learning-rate schedule, gives $\{h_i, \delta_j\}_D = c(t)(X\eta)_{ij}\,\delta(\tau - \tau')$. The kinetic term $\mathcal{K}_t/c(t)$ alone then adds $-(\dot c/c^2)\,\bm\delta$ to the error equation, which the algorithm does not have, and the Galley term $(\dot c/c^2)\,\mathbf{h}_-\cdot\bm\delta_+$, the analogue of the $\dot\gamma$ term of Ref.~\cite{holanda2026backpropagation}, removes it: it contains no $\bm\delta_-$, leaves the forward equation unchanged, and vanishes in the physical limit.

\emph{Unequal times.} Brackets between fields at different training times are not part of the canonical structure. The covariant bracket of Peierls~\cite{peierls1952commutation} defines them through the difference of the retarded and advanced Green's functions of the linearized equations; since the two pairings differ exactly by retarded and advanced error transport, it may give a covariant statement of their difference.

\bibliography{references}

%apsrev4-2.bst 2019-01-14 (MD) hand-edited version of apsrev4-1.bst
%Control: key (0)
%Control: author (8) initials jnrlst
%Control: editor formatted (1) identically to author
%Control: production of article title (0) allowed
%Control: page (0) single
%Control: year (1) truncated
%Control: production of eprint (0) enabled
\begin{thebibliography}{40}%
\makeatletter
\providecommand \@ifxundefined [1]{%
 \@ifx{#1\undefined}
}%
\providecommand \@ifnum [1]{%
 \ifnum #1\expandafter \@firstoftwo
 \else \expandafter \@secondoftwo
 \fi
}%
\providecommand \@ifx [1]{%
 \ifx #1\expandafter \@firstoftwo
 \else \expandafter \@secondoftwo
 \fi
}%
\providecommand \natexlab [1]{#1}%
\providecommand \enquote  [1]{``#1''}%
\providecommand \bibnamefont  [1]{#1}%
\providecommand \bibfnamefont [1]{#1}%
\providecommand \citenamefont [1]{#1}%
\providecommand \href@noop [0]{\@secondoftwo}%
\providecommand \href [0]{\begingroup \@sanitize@url \@href}%
\providecommand \@href[1]{\@@startlink{#1}\@@href}%
\providecommand \@@href[1]{\endgroup#1\@@endlink}%
\providecommand \@sanitize@url [0]{\catcode `\\12\catcode `\$12\catcode
  `\&12\catcode `\#12\catcode `\^12\catcode `\_12\catcode `\%12\relax}%
\providecommand \@@startlink[1]{}%
\providecommand \@@endlink[0]{}%
\providecommand \url  [0]{\begingroup\@sanitize@url \@url }%
\providecommand \@url [1]{\endgroup\@href {#1}{\urlprefix }}%
\providecommand \urlprefix  [0]{URL }%
\providecommand \Eprint [0]{\href }%
\providecommand \doibase [0]{https://doi.org/}%
\providecommand \selectlanguage [0]{\@gobble}%
\providecommand \bibinfo  [0]{\@secondoftwo}%
\providecommand \bibfield  [0]{\@secondoftwo}%
\providecommand \translation [1]{[#1]}%
\providecommand \BibitemOpen [0]{}%
\providecommand \bibitemStop [0]{}%
\providecommand \bibitemNoStop [0]{.\EOS\space}%
\providecommand \EOS [0]{\spacefactor3000\relax}%
\providecommand \BibitemShut  [1]{\csname bibitem#1\endcsname}%
\let\auto@bib@innerbib\@empty
%</preamble>
\bibitem [{\citenamefont {Rumelhart}\ \emph {et~al.}(1986)\citenamefont
  {Rumelhart}, \citenamefont {Hinton},\ and\ \citenamefont
  {Williams}}]{rumelhart1986learning}%
  \BibitemOpen
  \bibfield  {author} {\bibinfo {author} {\bibfnamefont {D.~E.}\ \bibnamefont
  {Rumelhart}}, \bibinfo {author} {\bibfnamefont {G.~E.}\ \bibnamefont
  {Hinton}},\ and\ \bibinfo {author} {\bibfnamefont {R.~J.}\ \bibnamefont
  {Williams}},\ }\bibfield  {title} {\bibinfo {title} {Learning representations
  by back-propagating errors},\ }\href {https://doi.org/10.1038/323533a0}
  {\bibfield  {journal} {\bibinfo  {journal} {Nature}\ }\textbf {\bibinfo
  {volume} {323}},\ \bibinfo {pages} {533} (\bibinfo {year}
  {1986})}\BibitemShut {NoStop}%
\bibitem [{\citenamefont {Chen}\ \emph
  {et~al.}(2018{\natexlab{a}})\citenamefont {Chen}, \citenamefont {Rubanova},
  \citenamefont {Bettencourt},\ and\ \citenamefont
  {Duvenaud}}]{chen2018neural}%
  \BibitemOpen
  \bibfield  {author} {\bibinfo {author} {\bibfnamefont {R.~T.~Q.}\
  \bibnamefont {Chen}}, \bibinfo {author} {\bibfnamefont {Y.}~\bibnamefont
  {Rubanova}}, \bibinfo {author} {\bibfnamefont {J.}~\bibnamefont
  {Bettencourt}},\ and\ \bibinfo {author} {\bibfnamefont {D.}~\bibnamefont
  {Duvenaud}},\ }\bibfield  {title} {\bibinfo {title} {Neural ordinary
  differential equations},\ }in\ \href@noop {} {\emph {\bibinfo {booktitle}
  {Advances in Neural Information Processing Systems}}},\ Vol.~\bibinfo
  {volume} {31}\ (\bibinfo {year} {2018})\BibitemShut {NoStop}%
\bibitem [{\citenamefont {LeCun}(1988)}]{lecun1988theoretical}%
  \BibitemOpen
  \bibfield  {author} {\bibinfo {author} {\bibfnamefont {Y.}~\bibnamefont
  {LeCun}},\ }\bibfield  {title} {\bibinfo {title} {A theoretical framework for
  back-propagation},\ }in\ \href@noop {} {\emph {\bibinfo {booktitle}
  {Proceedings of the 1988 Connectionist Models Summer School}}},\ \bibinfo
  {editor} {edited by\ \bibinfo {editor} {\bibfnamefont {D.}~\bibnamefont
  {Touretzky}}, \bibinfo {editor} {\bibfnamefont {G.}~\bibnamefont {Hinton}},\
  and\ \bibinfo {editor} {\bibfnamefont {T.}~\bibnamefont {Sejnowski}}}\
  (\bibinfo  {publisher} {Morgan Kaufmann},\ \bibinfo {address} {San Mateo,
  CA},\ \bibinfo {year} {1988})\ pp.\ \bibinfo {pages} {21--28}\BibitemShut
  {NoStop}%
\bibitem [{\citenamefont {E}\ \emph {et~al.}(2018)\citenamefont {E},
  \citenamefont {Han},\ and\ \citenamefont {Li}}]{e2018mean}%
  \BibitemOpen
  \bibfield  {author} {\bibinfo {author} {\bibfnamefont {W.}~\bibnamefont {E}},
  \bibinfo {author} {\bibfnamefont {J.}~\bibnamefont {Han}},\ and\ \bibinfo
  {author} {\bibfnamefont {Q.}~\bibnamefont {Li}},\ }\href@noop {} {\bibinfo
  {title} {A mean-field optimal control formulation of deep learning}}
  (\bibinfo {year} {2018}),\ \Eprint {https://arxiv.org/abs/1807.01083}
  {arXiv:1807.01083} \BibitemShut {NoStop}%
\bibitem [{\citenamefont {Scurria}(2026)}]{scurria2026physical}%
  \BibitemOpen
  \bibfield  {author} {\bibinfo {author} {\bibfnamefont {A.~E.}\ \bibnamefont
  {Scurria}},\ }\href@noop {} {\bibinfo {title} {A physical theory of
  backpropagation: Exact gradients from the least-action principle}} (\bibinfo
  {year} {2026}),\ \Eprint {https://arxiv.org/abs/2602.02281} {arXiv:2602.02281
  [cs.LG]} \BibitemShut {NoStop}%
\bibitem [{\citenamefont {Holanda}\ and\ \citenamefont
  {Cunha}(2026{\natexlab{a}})}]{holanda2026backpropagation}%
  \BibitemOpen
  \bibfield  {author} {\bibinfo {author} {\bibfnamefont {N.~L.}\ \bibnamefont
  {Holanda}}\ and\ \bibinfo {author} {\bibfnamefont {I.~E.}\ \bibnamefont
  {Cunha}},\ }\href@noop {} {\bibinfo {title} {Backpropagation as
  {Dirac}-bracket dynamics of dissipative training}} (\bibinfo {year}
  {2026}{\natexlab{a}}),\ \bibinfo {note} {submitted to Physical Review E},\
  \Eprint {https://arxiv.org/abs/2611.34479} {arXiv:2611.34479 [math-ph]}
  \BibitemShut {NoStop}%
\bibitem [{\citenamefont {Bateman}(1931)}]{bateman1931dissipative}%
  \BibitemOpen
  \bibfield  {author} {\bibinfo {author} {\bibfnamefont {H.}~\bibnamefont
  {Bateman}},\ }\bibfield  {title} {\bibinfo {title} {On dissipative systems
  and related variational principles},\ }\href@noop {} {\bibfield  {journal}
  {\bibinfo  {journal} {Physical Review}\ }\textbf {\bibinfo {volume} {38}},\
  \bibinfo {pages} {815} (\bibinfo {year} {1931})}\BibitemShut {NoStop}%
\bibitem [{\citenamefont {Galley}(2013)}]{galley2013classical}%
  \BibitemOpen
  \bibfield  {author} {\bibinfo {author} {\bibfnamefont {C.~R.}\ \bibnamefont
  {Galley}},\ }\bibfield  {title} {\bibinfo {title} {Classical mechanics of
  nonconservative systems},\ }\href@noop {} {\bibfield  {journal} {\bibinfo
  {journal} {Physical review letters}\ }\textbf {\bibinfo {volume} {110}},\
  \bibinfo {pages} {174301} (\bibinfo {year} {2013})}\BibitemShut {NoStop}%
\bibitem [{\citenamefont {Galley}\ \emph {et~al.}(2014)\citenamefont {Galley},
  \citenamefont {Tsang},\ and\ \citenamefont {Stein}}]{galley2014principle}%
  \BibitemOpen
  \bibfield  {author} {\bibinfo {author} {\bibfnamefont {C.~R.}\ \bibnamefont
  {Galley}}, \bibinfo {author} {\bibfnamefont {D.}~\bibnamefont {Tsang}},\ and\
  \bibinfo {author} {\bibfnamefont {L.~C.}\ \bibnamefont {Stein}},\ }\bibfield
  {title} {\bibinfo {title} {The principle of stationary nonconservative action
  for classical mechanics and field theories},\ }\href@noop {} {\bibfield
  {journal} {\bibinfo  {journal} {arXiv preprint arXiv:1412.3082}\ } (\bibinfo
  {year} {2014})}\BibitemShut {NoStop}%
\bibitem [{\citenamefont {Dirac}(1950)}]{dirac1950generalized}%
  \BibitemOpen
  \bibfield  {author} {\bibinfo {author} {\bibfnamefont {P.~A.~M.}\
  \bibnamefont {Dirac}},\ }\bibfield  {title} {\bibinfo {title} {Generalized
  hamiltonian dynamics},\ }\href@noop {} {\bibfield  {journal} {\bibinfo
  {journal} {Canadian journal of mathematics}\ }\textbf {\bibinfo {volume}
  {2}},\ \bibinfo {pages} {129} (\bibinfo {year} {1950})}\BibitemShut {NoStop}%
\bibitem [{\citenamefont {Henneaux}\ and\ \citenamefont
  {Teitelboim}(1992)}]{henneaux1992quantization}%
  \BibitemOpen
  \bibfield  {author} {\bibinfo {author} {\bibfnamefont {M.}~\bibnamefont
  {Henneaux}}\ and\ \bibinfo {author} {\bibfnamefont {C.}~\bibnamefont
  {Teitelboim}},\ }\href@noop {} {\emph {\bibinfo {title} {Quantization of
  Gauge Systems}}}\ (\bibinfo  {publisher} {Princeton University Press},\
  \bibinfo {address} {Princeton, NJ},\ \bibinfo {year} {1992})\BibitemShut
  {NoStop}%
\bibitem [{\citenamefont {Lu}\ \emph {et~al.}(2020)\citenamefont {Lu},
  \citenamefont {Ma}, \citenamefont {Lu}, \citenamefont {Lu},\ and\
  \citenamefont {Ying}}]{lu2020mean}%
  \BibitemOpen
  \bibfield  {author} {\bibinfo {author} {\bibfnamefont {Y.}~\bibnamefont
  {Lu}}, \bibinfo {author} {\bibfnamefont {C.}~\bibnamefont {Ma}}, \bibinfo
  {author} {\bibfnamefont {Y.}~\bibnamefont {Lu}}, \bibinfo {author}
  {\bibfnamefont {J.}~\bibnamefont {Lu}},\ and\ \bibinfo {author}
  {\bibfnamefont {L.}~\bibnamefont {Ying}},\ }\href
  {https://doi.org/10.48550/arXiv.2003.05508} {\bibinfo {title} {A mean-field
  analysis of deep {ResNet} and beyond: Towards provable optimization via
  overparameterization from depth}} (\bibinfo {year} {2020}),\ \Eprint
  {https://arxiv.org/abs/2003.05508} {arXiv:2003.05508} \BibitemShut {NoStop}%
\bibitem [{\citenamefont {Marion}\ \emph {et~al.}(2023)\citenamefont {Marion},
  \citenamefont {Wu}, \citenamefont {Sander},\ and\ \citenamefont
  {Biau}}]{marion2023implicit}%
  \BibitemOpen
  \bibfield  {author} {\bibinfo {author} {\bibfnamefont {P.}~\bibnamefont
  {Marion}}, \bibinfo {author} {\bibfnamefont {Y.-H.}\ \bibnamefont {Wu}},
  \bibinfo {author} {\bibfnamefont {M.~E.}\ \bibnamefont {Sander}},\ and\
  \bibinfo {author} {\bibfnamefont {G.}~\bibnamefont {Biau}},\ }\href
  {https://doi.org/10.48550/arXiv.2309.01213} {\bibinfo {title} {Implicit
  regularization of deep residual networks towards neural {ODEs}}} (\bibinfo
  {year} {2023}),\ \bibinfo {note} {iCLR 2024},\ \Eprint
  {https://arxiv.org/abs/2309.01213} {arXiv:2309.01213} \BibitemShut {NoStop}%
\bibitem [{\citenamefont {Lee}\ \emph {et~al.}(2025)\citenamefont {Lee},
  \citenamefont {Lee},\ and\ \citenamefont {Yi}}]{lee2025synaptic}%
  \BibitemOpen
  \bibfield  {author} {\bibinfo {author} {\bibfnamefont {D.}~\bibnamefont
  {Lee}}, \bibinfo {author} {\bibfnamefont {H.-S.}\ \bibnamefont {Lee}},\ and\
  \bibinfo {author} {\bibfnamefont {J.}~\bibnamefont {Yi}},\ }\href@noop {}
  {\bibinfo {title} {Synaptic field theory for neural networks}} (\bibinfo
  {year} {2025}),\ \Eprint {https://arxiv.org/abs/2503.08827} {arXiv:2503.08827
  [hep-th]} \BibitemShut {NoStop}%
\bibitem [{\citenamefont {Pehle}\ and\ \citenamefont
  {Slotine}(2026)}]{pehle2026unlocked}%
  \BibitemOpen
  \bibfield  {author} {\bibinfo {author} {\bibfnamefont {C.}~\bibnamefont
  {Pehle}}\ and\ \bibinfo {author} {\bibfnamefont {J.-J.}\ \bibnamefont
  {Slotine}},\ }\href {https://doi.org/10.48550/arXiv.2602.10461} {\bibinfo
  {title} {Unlocked backpropagation using wave scattering}} (\bibinfo {year}
  {2026}),\ \Eprint {https://arxiv.org/abs/2602.10461} {arXiv:2602.10461}
  \BibitemShut {NoStop}%
\bibitem [{\citenamefont {Ellenberger}\ \emph {et~al.}(2026)\citenamefont
  {Ellenberger}, \citenamefont {Haider}, \citenamefont {Jordan}, \citenamefont
  {Max}, \citenamefont {Jaras}, \citenamefont {Kriener}, \citenamefont
  {Benitez},\ and\ \citenamefont {Petrovici}}]{ellenberger2026backpropagation}%
  \BibitemOpen
  \bibfield  {author} {\bibinfo {author} {\bibfnamefont {B.}~\bibnamefont
  {Ellenberger}}, \bibinfo {author} {\bibfnamefont {P.}~\bibnamefont {Haider}},
  \bibinfo {author} {\bibfnamefont {J.}~\bibnamefont {Jordan}}, \bibinfo
  {author} {\bibfnamefont {K.}~\bibnamefont {Max}}, \bibinfo {author}
  {\bibfnamefont {I.}~\bibnamefont {Jaras}}, \bibinfo {author} {\bibfnamefont
  {L.}~\bibnamefont {Kriener}}, \bibinfo {author} {\bibfnamefont
  {F.}~\bibnamefont {Benitez}},\ and\ \bibinfo {author} {\bibfnamefont {M.~A.}\
  \bibnamefont {Petrovici}},\ }\bibfield  {title} {\bibinfo {title}
  {Backpropagation through space, time, and the brain},\ }\href
  {https://doi.org/10.1038/s41467-025-66666-z} {\bibfield  {journal} {\bibinfo
  {journal} {Nat. Commun.}\ }\textbf {\bibinfo {volume} {17}},\ \bibinfo
  {pages} {66} (\bibinfo {year} {2026})}\BibitemShut {NoStop}%
\bibitem [{\citenamefont {Mitliagkas}\ \emph {et~al.}(2016)\citenamefont
  {Mitliagkas}, \citenamefont {Zhang}, \citenamefont {Hadjis},\ and\
  \citenamefont {R{\'e}}}]{mitliagkas2016asynchrony}%
  \BibitemOpen
  \bibfield  {author} {\bibinfo {author} {\bibfnamefont {I.}~\bibnamefont
  {Mitliagkas}}, \bibinfo {author} {\bibfnamefont {C.}~\bibnamefont {Zhang}},
  \bibinfo {author} {\bibfnamefont {S.}~\bibnamefont {Hadjis}},\ and\ \bibinfo
  {author} {\bibfnamefont {C.}~\bibnamefont {R{\'e}}},\ }\href
  {https://doi.org/10.48550/arXiv.1605.09774} {\bibinfo {title} {Asynchrony
  begets momentum, with an application to deep learning}} (\bibinfo {year}
  {2016}),\ \Eprint {https://arxiv.org/abs/1605.09774} {arXiv:1605.09774}
  \BibitemShut {NoStop}%
\bibitem [{\citenamefont {Polyak}(1964)}]{polyak1964some}%
  \BibitemOpen
  \bibfield  {author} {\bibinfo {author} {\bibfnamefont {B.~T.}\ \bibnamefont
  {Polyak}},\ }\bibfield  {title} {\bibinfo {title} {Some methods of speeding
  up the convergence of iteration methods},\ }\href
  {https://doi.org/10.1016/0041-5553(64)90137-5} {\bibfield  {journal}
  {\bibinfo  {journal} {USSR Computational Mathematics and Mathematical
  Physics}\ }\textbf {\bibinfo {volume} {4}},\ \bibinfo {pages} {1} (\bibinfo
  {year} {1964})}\BibitemShut {NoStop}%
\bibitem [{\citenamefont {Nesterov}(1983)}]{nesterov1983method}%
  \BibitemOpen
  \bibfield  {author} {\bibinfo {author} {\bibfnamefont {Y.~E.}\ \bibnamefont
  {Nesterov}},\ }\bibfield  {title} {\bibinfo {title} {A method of solving a
  convex programming problem with convergence rate {$O(1/k^2)$}},\ }\href@noop
  {} {\bibfield  {journal} {\bibinfo  {journal} {Doklady Akademii Nauk SSSR}\
  }\textbf {\bibinfo {volume} {269}},\ \bibinfo {pages} {543} (\bibinfo {year}
  {1983})}\BibitemShut {NoStop}%
\bibitem [{\citenamefont {Chen}\ \emph
  {et~al.}(2018{\natexlab{b}})\citenamefont {Chen}, \citenamefont {Yang},\ and\
  \citenamefont {Cheng}}]{chen2018efficient}%
  \BibitemOpen
  \bibfield  {author} {\bibinfo {author} {\bibfnamefont {C.-C.}\ \bibnamefont
  {Chen}}, \bibinfo {author} {\bibfnamefont {C.-L.}\ \bibnamefont {Yang}},\
  and\ \bibinfo {author} {\bibfnamefont {H.-Y.}\ \bibnamefont {Cheng}},\ }\href
  {https://doi.org/10.48550/arXiv.1809.02839} {\bibinfo {title} {Efficient and
  robust parallel {DNN} training through model parallelism on multi-{GPU}
  platform}} (\bibinfo {year} {2018}{\natexlab{b}}),\ \Eprint
  {https://arxiv.org/abs/1809.02839} {arXiv:1809.02839} \BibitemShut {NoStop}%
\bibitem [{\citenamefont {Kosson}\ \emph {et~al.}(2020)\citenamefont {Kosson},
  \citenamefont {Chiley}, \citenamefont {Venigalla}, \citenamefont {Hestness},\
  and\ \citenamefont {K{\"o}ster}}]{kosson2020pipelined}%
  \BibitemOpen
  \bibfield  {author} {\bibinfo {author} {\bibfnamefont {A.}~\bibnamefont
  {Kosson}}, \bibinfo {author} {\bibfnamefont {V.}~\bibnamefont {Chiley}},
  \bibinfo {author} {\bibfnamefont {A.}~\bibnamefont {Venigalla}}, \bibinfo
  {author} {\bibfnamefont {J.}~\bibnamefont {Hestness}},\ and\ \bibinfo
  {author} {\bibfnamefont {U.}~\bibnamefont {K{\"o}ster}},\ }\href
  {https://doi.org/10.48550/arXiv.2003.11666} {\bibinfo {title} {Pipelined
  backpropagation at scale: Training large models without batches}} (\bibinfo
  {year} {2020}),\ \bibinfo {note} {proceedings of the 4th MLSys Conference,
  2021},\ \Eprint {https://arxiv.org/abs/2003.11666} {arXiv:2003.11666}
  \BibitemShut {NoStop}%
\bibitem [{\citenamefont {Jacot}\ \emph {et~al.}(2018)\citenamefont {Jacot},
  \citenamefont {Gabriel},\ and\ \citenamefont {Hongler}}]{jacot2018neural}%
  \BibitemOpen
  \bibfield  {author} {\bibinfo {author} {\bibfnamefont {A.}~\bibnamefont
  {Jacot}}, \bibinfo {author} {\bibfnamefont {F.}~\bibnamefont {Gabriel}},\
  and\ \bibinfo {author} {\bibfnamefont {C.}~\bibnamefont {Hongler}},\
  }\bibfield  {title} {\bibinfo {title} {Neural tangent kernel: Convergence and
  generalization in neural networks},\ }in\ \href@noop {} {\emph {\bibinfo
  {booktitle} {Advances in Neural Information Processing Systems}}},\
  Vol.~\bibinfo {volume} {31}\ (\bibinfo {year} {2018})\BibitemShut {NoStop}%
\bibitem [{\citenamefont {Janson}\ and\ \citenamefont
  {Marsden}(2021)}]{janson2021delay}%
  \BibitemOpen
  \bibfield  {author} {\bibinfo {author} {\bibfnamefont {N.~B.}\ \bibnamefont
  {Janson}}\ and\ \bibinfo {author} {\bibfnamefont {C.~J.}\ \bibnamefont
  {Marsden}},\ }\bibfield  {title} {\bibinfo {title} {Delay-induced homoclinic
  bifurcations in modified gradient bistable systems and their relevance to
  optimisation},\ }\href {https://doi.org/10.1063/5.0035959} {\bibfield
  {journal} {\bibinfo  {journal} {Chaos}\ }\textbf {\bibinfo {volume} {31}},\
  \bibinfo {pages} {093120} (\bibinfo {year} {2021})}\BibitemShut {NoStop}%
\bibitem [{\citenamefont {Giladi}\ \emph {et~al.}(2019)\citenamefont {Giladi},
  \citenamefont {Shpigel~Nacson}, \citenamefont {Hoffer},\ and\ \citenamefont
  {Soudry}}]{giladi2019stability}%
  \BibitemOpen
  \bibfield  {author} {\bibinfo {author} {\bibfnamefont {N.}~\bibnamefont
  {Giladi}}, \bibinfo {author} {\bibfnamefont {M.}~\bibnamefont
  {Shpigel~Nacson}}, \bibinfo {author} {\bibfnamefont {E.}~\bibnamefont
  {Hoffer}},\ and\ \bibinfo {author} {\bibfnamefont {D.}~\bibnamefont
  {Soudry}},\ }\href {https://doi.org/10.48550/arXiv.1909.12340} {\bibinfo
  {title} {At stability's edge: How to adjust hyperparameters to preserve
  minima selection in asynchronous training of neural networks?}} (\bibinfo
  {year} {2019}),\ \bibinfo {note} {iCLR 2020},\ \Eprint
  {https://arxiv.org/abs/1909.12340} {arXiv:1909.12340} \BibitemShut {NoStop}%
\bibitem [{\citenamefont {Defazio}\ \emph {et~al.}(2024)\citenamefont
  {Defazio}, \citenamefont {Yang}, \citenamefont {Mehta}, \citenamefont
  {Mishchenko}, \citenamefont {Khaled},\ and\ \citenamefont
  {Cutkosky}}]{defazio2024road}%
  \BibitemOpen
  \bibfield  {author} {\bibinfo {author} {\bibfnamefont {A.}~\bibnamefont
  {Defazio}}, \bibinfo {author} {\bibfnamefont {X.}~\bibnamefont {Yang}},
  \bibinfo {author} {\bibfnamefont {H.}~\bibnamefont {Mehta}}, \bibinfo
  {author} {\bibfnamefont {K.}~\bibnamefont {Mishchenko}}, \bibinfo {author}
  {\bibfnamefont {A.}~\bibnamefont {Khaled}},\ and\ \bibinfo {author}
  {\bibfnamefont {A.}~\bibnamefont {Cutkosky}},\ }\bibfield  {title} {\bibinfo
  {title} {The road less scheduled},\ }in\ \href
  {https://doi.org/10.52202/079017-0320} {\emph {\bibinfo {booktitle} {Advances
  in Neural Information Processing Systems}}},\ Vol.~\bibinfo {volume} {37},\
  \bibinfo {editor} {edited by\ \bibinfo {editor} {\bibfnamefont
  {A.}~\bibnamefont {Globerson}}, \bibinfo {editor} {\bibfnamefont
  {L.}~\bibnamefont {Mackey}}, \bibinfo {editor} {\bibfnamefont
  {D.}~\bibnamefont {Belgrave}}, \bibinfo {editor} {\bibfnamefont
  {A.}~\bibnamefont {Fan}}, \bibinfo {editor} {\bibfnamefont {U.}~\bibnamefont
  {Paquet}}, \bibinfo {editor} {\bibfnamefont {J.}~\bibnamefont {Tomczak}},\
  and\ \bibinfo {editor} {\bibfnamefont {C.}~\bibnamefont {Zhang}}}\ (\bibinfo
  {publisher} {Curran Associates, Inc.},\ \bibinfo {year} {2024})\ pp.\
  \bibinfo {pages} {9974--10007},\ \Eprint {https://arxiv.org/abs/2405.15682}
  {arXiv:2405.15682} \BibitemShut {NoStop}%
\bibitem [{\citenamefont {Morwani}\ \emph {et~al.}(2025)\citenamefont
  {Morwani}, \citenamefont {Vyas}, \citenamefont {Zhang},\ and\ \citenamefont
  {Kakade}}]{morwani2025connections}%
  \BibitemOpen
  \bibfield  {author} {\bibinfo {author} {\bibfnamefont {D.}~\bibnamefont
  {Morwani}}, \bibinfo {author} {\bibfnamefont {N.}~\bibnamefont {Vyas}},
  \bibinfo {author} {\bibfnamefont {H.}~\bibnamefont {Zhang}},\ and\ \bibinfo
  {author} {\bibfnamefont {S.}~\bibnamefont {Kakade}},\ }\href
  {https://doi.org/10.48550/arXiv.2502.02431} {\bibinfo {title} {Connections
  between schedule-free optimizers, {AdEMAMix}, and accelerated {SGD}
  variants}} (\bibinfo {year} {2025}),\ \Eprint
  {https://arxiv.org/abs/2502.02431} {arXiv:2502.02431} \BibitemShut {NoStop}%
\bibitem [{\citenamefont {Holanda}\ and\ \citenamefont
  {Cunha}(2026{\natexlab{b}})}]{holanda2026twotime}%
  \BibitemOpen
  \bibfield  {author} {\bibinfo {author} {\bibfnamefont {N.~L.}\ \bibnamefont
  {Holanda}}\ and\ \bibinfo {author} {\bibfnamefont {I.~E.}\ \bibnamefont
  {Cunha}},\ }\href@noop {} {\bibinfo {title} {{two-time-action-code}:
  numerical tests for ``{Training time and depth as a $1+1$ field theory of
  learning}''}},\ \bibinfo {howpublished}
  {\url{https://github.com/linneuholanda/two-time-action-code}} (\bibinfo
  {year} {2026}{\natexlab{b}})\BibitemShut {NoStop}%
\bibitem [{\citenamefont {Yang}\ \emph {et~al.}(2023)\citenamefont {Yang},
  \citenamefont {Yu}, \citenamefont {Zhu},\ and\ \citenamefont
  {Hayou}}]{yang2023tensor}%
  \BibitemOpen
  \bibfield  {author} {\bibinfo {author} {\bibfnamefont {G.}~\bibnamefont
  {Yang}}, \bibinfo {author} {\bibfnamefont {D.}~\bibnamefont {Yu}}, \bibinfo
  {author} {\bibfnamefont {C.}~\bibnamefont {Zhu}},\ and\ \bibinfo {author}
  {\bibfnamefont {S.}~\bibnamefont {Hayou}},\ }\href
  {https://doi.org/10.48550/arXiv.2310.02244} {\bibinfo {title} {Tensor
  programs {VI}: Feature learning in infinite-depth neural networks}} (\bibinfo
  {year} {2023}),\ \Eprint {https://arxiv.org/abs/2310.02244}
  {arXiv:2310.02244} \BibitemShut {NoStop}%
\bibitem [{\citenamefont {Bordelon}\ \emph {et~al.}(2023)\citenamefont
  {Bordelon}, \citenamefont {Noci}, \citenamefont {Li}, \citenamefont {Hanin},\
  and\ \citenamefont {Pehlevan}}]{bordelon2023depthwise}%
  \BibitemOpen
  \bibfield  {author} {\bibinfo {author} {\bibfnamefont {B.}~\bibnamefont
  {Bordelon}}, \bibinfo {author} {\bibfnamefont {L.}~\bibnamefont {Noci}},
  \bibinfo {author} {\bibfnamefont {M.~B.}\ \bibnamefont {Li}}, \bibinfo
  {author} {\bibfnamefont {B.}~\bibnamefont {Hanin}},\ and\ \bibinfo {author}
  {\bibfnamefont {C.}~\bibnamefont {Pehlevan}},\ }\href
  {https://doi.org/10.48550/arXiv.2309.16620} {\bibinfo {title} {Depthwise
  hyperparameter transfer in residual networks: Dynamics and scaling limit}}
  (\bibinfo {year} {2023}),\ \Eprint {https://arxiv.org/abs/2309.16620}
  {arXiv:2309.16620} \BibitemShut {NoStop}%
\bibitem [{\citenamefont {Gitman}\ and\ \citenamefont
  {Tyutin}(1990)}]{gitman1990quantization}%
  \BibitemOpen
  \bibfield  {author} {\bibinfo {author} {\bibfnamefont {D.~M.}\ \bibnamefont
  {Gitman}}\ and\ \bibinfo {author} {\bibfnamefont {I.~V.}\ \bibnamefont
  {Tyutin}},\ }\href {https://doi.org/10.1007/978-3-642-83938-2} {\emph
  {\bibinfo {title} {Quantization of Fields with Constraints}}},\ Springer
  Series in Nuclear and Particle Physics\ (\bibinfo  {publisher} {Springer},\
  \bibinfo {address} {Berlin, Heidelberg},\ \bibinfo {year} {1990})\BibitemShut
  {NoStop}%
\bibitem [{\citenamefont {Harlap}\ \emph {et~al.}(2018)\citenamefont {Harlap},
  \citenamefont {Narayanan}, \citenamefont {Phanishayee}, \citenamefont
  {Seshadri}, \citenamefont {Devanur}, \citenamefont {Ganger},\ and\
  \citenamefont {Gibbons}}]{harlap2018pipedream}%
  \BibitemOpen
  \bibfield  {author} {\bibinfo {author} {\bibfnamefont {A.}~\bibnamefont
  {Harlap}}, \bibinfo {author} {\bibfnamefont {D.}~\bibnamefont {Narayanan}},
  \bibinfo {author} {\bibfnamefont {A.}~\bibnamefont {Phanishayee}}, \bibinfo
  {author} {\bibfnamefont {V.}~\bibnamefont {Seshadri}}, \bibinfo {author}
  {\bibfnamefont {N.}~\bibnamefont {Devanur}}, \bibinfo {author} {\bibfnamefont
  {G.}~\bibnamefont {Ganger}},\ and\ \bibinfo {author} {\bibfnamefont
  {P.}~\bibnamefont {Gibbons}},\ }\href
  {https://doi.org/10.48550/arXiv.1806.03377} {\bibinfo {title} {{PipeDream}:
  Fast and efficient pipeline parallel {DNN} training}} (\bibinfo {year}
  {2018}),\ \Eprint {https://arxiv.org/abs/1806.03377} {arXiv:1806.03377}
  \BibitemShut {NoStop}%
\bibitem [{\citenamefont {Yang}\ \emph {et~al.}(2019)\citenamefont {Yang},
  \citenamefont {Zhang}, \citenamefont {Li}, \citenamefont {R{\'e}},
  \citenamefont {Aberger},\ and\ \citenamefont {De~Sa}}]{yang2019pipemare}%
  \BibitemOpen
  \bibfield  {author} {\bibinfo {author} {\bibfnamefont {B.}~\bibnamefont
  {Yang}}, \bibinfo {author} {\bibfnamefont {J.}~\bibnamefont {Zhang}},
  \bibinfo {author} {\bibfnamefont {J.}~\bibnamefont {Li}}, \bibinfo {author}
  {\bibfnamefont {C.}~\bibnamefont {R{\'e}}}, \bibinfo {author} {\bibfnamefont
  {C.~R.}\ \bibnamefont {Aberger}},\ and\ \bibinfo {author} {\bibfnamefont
  {C.}~\bibnamefont {De~Sa}},\ }\href
  {https://doi.org/10.48550/arXiv.1910.05124} {\bibinfo {title} {{PipeMare}:
  Asynchronous pipeline parallel {DNN} training}} (\bibinfo {year} {2019}),\
  \Eprint {https://arxiv.org/abs/1910.05124} {arXiv:1910.05124} \BibitemShut
  {NoStop}%
\bibitem [{\citenamefont {Stich}\ and\ \citenamefont
  {Karimireddy}(2020)}]{stich2020error}%
  \BibitemOpen
  \bibfield  {author} {\bibinfo {author} {\bibfnamefont {S.~U.}\ \bibnamefont
  {Stich}}\ and\ \bibinfo {author} {\bibfnamefont {S.~P.}\ \bibnamefont
  {Karimireddy}},\ }\bibfield  {title} {\bibinfo {title} {The error-feedback
  framework: Better rates for {SGD} with delayed gradients and compressed
  communication},\ }\href@noop {} {\bibfield  {journal} {\bibinfo  {journal}
  {Journal of Machine Learning Research}\ }\textbf {\bibinfo {volume} {21}},\
  \bibinfo {pages} {1} (\bibinfo {year} {2020})}\BibitemShut {NoStop}%
\bibitem [{\citenamefont {Zheng}\ \emph {et~al.}(2016)\citenamefont {Zheng},
  \citenamefont {Meng}, \citenamefont {Wang}, \citenamefont {Chen},
  \citenamefont {Yu}, \citenamefont {Ma},\ and\ \citenamefont
  {Liu}}]{zheng2017asynchronous}%
  \BibitemOpen
  \bibfield  {author} {\bibinfo {author} {\bibfnamefont {S.}~\bibnamefont
  {Zheng}}, \bibinfo {author} {\bibfnamefont {Q.}~\bibnamefont {Meng}},
  \bibinfo {author} {\bibfnamefont {T.}~\bibnamefont {Wang}}, \bibinfo {author}
  {\bibfnamefont {W.}~\bibnamefont {Chen}}, \bibinfo {author} {\bibfnamefont
  {N.}~\bibnamefont {Yu}}, \bibinfo {author} {\bibfnamefont {Z.-M.}\
  \bibnamefont {Ma}},\ and\ \bibinfo {author} {\bibfnamefont {T.-Y.}\
  \bibnamefont {Liu}},\ }\href {https://doi.org/10.48550/arXiv.1609.08326}
  {\bibinfo {title} {Asynchronous stochastic gradient descent with delay
  compensation}} (\bibinfo {year} {2016}),\ \bibinfo {note} {international
  Conference on Machine Learning 2017, pp. 4120--4129},\ \Eprint
  {https://arxiv.org/abs/1609.08326} {arXiv:1609.08326} \BibitemShut {NoStop}%
\bibitem [{\citenamefont {Du}\ \emph {et~al.}(2018)\citenamefont {Du},
  \citenamefont {Hu},\ and\ \citenamefont {Lee}}]{du2018algorithmic}%
  \BibitemOpen
  \bibfield  {author} {\bibinfo {author} {\bibfnamefont {S.~S.}\ \bibnamefont
  {Du}}, \bibinfo {author} {\bibfnamefont {W.}~\bibnamefont {Hu}},\ and\
  \bibinfo {author} {\bibfnamefont {J.~D.}\ \bibnamefont {Lee}},\ }\href
  {https://doi.org/10.48550/arXiv.1806.00900} {\bibinfo {title} {Algorithmic
  regularization in learning deep homogeneous models: Layers are automatically
  balanced}} (\bibinfo {year} {2018}),\ \Eprint
  {https://arxiv.org/abs/1806.00900} {arXiv:1806.00900} \BibitemShut {NoStop}%
\bibitem [{\citenamefont {Kunin}\ \emph {et~al.}(2020)\citenamefont {Kunin},
  \citenamefont {Sagastuy-Brena}, \citenamefont {Ganguli}, \citenamefont
  {Yamins},\ and\ \citenamefont {Tanaka}}]{kunin2020neural}%
  \BibitemOpen
  \bibfield  {author} {\bibinfo {author} {\bibfnamefont {D.}~\bibnamefont
  {Kunin}}, \bibinfo {author} {\bibfnamefont {J.}~\bibnamefont
  {Sagastuy-Brena}}, \bibinfo {author} {\bibfnamefont {S.}~\bibnamefont
  {Ganguli}}, \bibinfo {author} {\bibfnamefont {D.~L.}\ \bibnamefont
  {Yamins}},\ and\ \bibinfo {author} {\bibfnamefont {H.}~\bibnamefont
  {Tanaka}},\ }\bibfield  {title} {\bibinfo {title} {Neural mechanics: Symmetry
  and broken conservation laws in deep learning dynamics},\ }\href@noop {}
  {\bibfield  {journal} {\bibinfo  {journal} {arXiv preprint arXiv:2012.04728}\
  } (\bibinfo {year} {2020})}\BibitemShut {NoStop}%
\bibitem [{\citenamefont {Scellier}\ and\ \citenamefont
  {Bengio}(2016)}]{scellier2016equilibrium}%
  \BibitemOpen
  \bibfield  {author} {\bibinfo {author} {\bibfnamefont {B.}~\bibnamefont
  {Scellier}}\ and\ \bibinfo {author} {\bibfnamefont {Y.}~\bibnamefont
  {Bengio}},\ }\href {https://doi.org/10.48550/arXiv.1602.05179} {\bibinfo
  {title} {Equilibrium propagation: Bridging the gap between energy-based
  models and backpropagation}} (\bibinfo {year} {2016}),\ \Eprint
  {https://arxiv.org/abs/1602.05179} {arXiv:1602.05179} \BibitemShut {NoStop}%
\bibitem [{\citenamefont {Rao}\ and\ \citenamefont
  {Ballard}(1999)}]{rao1999predictive}%
  \BibitemOpen
  \bibfield  {author} {\bibinfo {author} {\bibfnamefont {R.~P.~N.}\
  \bibnamefont {Rao}}\ and\ \bibinfo {author} {\bibfnamefont {D.~H.}\
  \bibnamefont {Ballard}},\ }\bibfield  {title} {\bibinfo {title} {Predictive
  coding in the visual cortex: a functional interpretation of some
  extra-classical receptive-field effects},\ }\href
  {https://doi.org/10.1038/4580} {\bibfield  {journal} {\bibinfo  {journal}
  {Nature Neuroscience}\ }\textbf {\bibinfo {volume} {2}},\ \bibinfo {pages}
  {79} (\bibinfo {year} {1999})}\BibitemShut {NoStop}%
\bibitem [{\citenamefont {Whittington}\ and\ \citenamefont
  {Bogacz}(2017)}]{whittington2017approximation}%
  \BibitemOpen
  \bibfield  {author} {\bibinfo {author} {\bibfnamefont {J.~C.~R.}\
  \bibnamefont {Whittington}}\ and\ \bibinfo {author} {\bibfnamefont
  {R.}~\bibnamefont {Bogacz}},\ }\bibfield  {title} {\bibinfo {title} {An
  approximation of the error backpropagation algorithm in a predictive coding
  network with local {Hebbian} synaptic plasticity},\ }\href
  {https://doi.org/10.1162/NECO_a_00949} {\bibfield  {journal} {\bibinfo
  {journal} {Neural Computation}\ }\textbf {\bibinfo {volume} {29}},\ \bibinfo
  {pages} {1229} (\bibinfo {year} {2017})}\BibitemShut {NoStop}%
\bibitem [{\citenamefont {Peierls}(1952)}]{peierls1952commutation}%
  \BibitemOpen
  \bibfield  {author} {\bibinfo {author} {\bibfnamefont {R.~E.}\ \bibnamefont
  {Peierls}},\ }\bibfield  {title} {\bibinfo {title} {The commutation laws of
  relativistic field theory},\ }\href {https://doi.org/10.1098/rspa.1952.0158}
  {\bibfield  {journal} {\bibinfo  {journal} {Proceedings of the Royal Society
  of London. Series A. Mathematical and Physical Sciences}\ }\textbf {\bibinfo
  {volume} {214}},\ \bibinfo {pages} {143} (\bibinfo {year}
  {1952})}\BibitemShut {NoStop}%
\end{thebibliography}%

\end{document}